\documentclass[11pt]{article}

\usepackage[preprint]{acl}

\usepackage{times}
\usepackage{latexsym}
\usepackage{makecell}

\usepackage{booktabs}       
\usepackage{amsfonts}       
\usepackage{nicefrac}       
\usepackage{microtype}      
\usepackage{comment}
\usepackage{amsmath} 

\usepackage{algorithm}
\usepackage{algpseudocode}
\usepackage{graphicx}

\usepackage{multirow}
\usepackage{multicol}
\usepackage{booktabs}
\usepackage{caption}
\usepackage{makecell}
\usepackage{subcaption}
\usepackage{pifont}
\usepackage{wrapfig}
\usepackage{natbib}
\usepackage{amssymb} 
\usepackage{tabularx}
\usepackage[utf8]{inputenc}
\usepackage{newunicodechar}
\newunicodechar{⇔}{$\Leftrightarrow$}

\usepackage[T1]{fontenc}

\usepackage[utf8]{inputenc}

\usepackage{microtype}

\usepackage{inconsolata}

\usepackage{graphicx}

\newcommand{\cmark}{\ding{51}}
\newcommand{\xmark}{\ding{55}}

\title{\textsc{FOL-Traces}:\\ Verified 
First-Order Logic Reasoning Traces at Scale}

\author{Isabelle Lee \\
  USC \\
  \texttt{lee.isabelle.g@gmail.com} \\\And
  Sarah Liaw \\
  Harvard University \\ \\\And
  Dani Yogatama \\
  USC \\}

\begin{document}
\maketitle
\begin{abstract}

Reasoning in language models is difficult to evaluate: natural-language traces are unverifiable, symbolic datasets are too small, and most benchmarks conflate heuristics with inference. 
We present \textsc{FOL-Traces}, the first large-scale dataset of programmatically verified reasoning traces, enabling rigorous evaluation of structured logical inference.\footnote{The dataset is available at \url{https://huggingface.co/datasets/fol-traces/fol-traces}, and the generation and analysis code is available at \url{https://github.com/iglee/fol-traces}}
We also propose two challenging and comprehensive diagnostic tasks---\textit{masked operation prediction} and \textit{step completion}---that directly probe syntactic awareness and process fidelity.
\textsc{FOL-Traces} serves as a scalable testbed for rigorously studying how models perform structured logical inference.
Systematic experiments with 5 reasoning LLMs show that the dataset remains challenging: models only reach around 45.7\% accuracy on masked operation prediction and around 27\% on two-step completion.

\end{abstract}

\section{Introduction}

To establish whether large language models (LLMs) produce correct reasoning, they must be evaluated in domains that provide both structural clarity and verifiable inference. Existing approaches fall short of this requirement. Natural-language chains of thought, while human-interpretable, are poorly defined \citep{bender-koller-2020-climbing}, unverifiable, and often unfaithful to the model’s underlying computation \cite{tyen2024llmsreasoningerrorscorrect, min2022rethinkingroledemonstrationsmakes}. Symbolic datasets such as math problem sets, algebraic puzzles, and theorem-proving corpora provide formal supervision, but they are typically too small or under-annotated for large-scale, systematic reasoning analysis. 
Widely used benchmarks further conflate heuristic shortcuts with inference, making it difficult to determine whether success reflects systematic reasoning or pattern exploitation.

Recent advances in reasoning-augmented language models---such as OpenAI’s O1 \citep{openai2024o1}, DeepSeek’s R1 \citep{deepseekai2025deepseekr1incentivizingreasoningcapability}, and the Qwen reasoning model \citep{bai2023qwentechnicalreport}---have intensified interest in structured problem-solving within LLMs. 
These models have shown strong performance on complex tasks, including AIME mathematical challenges \citep{vals2025aime} and coding benchmarks \citep{chen2021codex, jimenez2024swebench}, often leveraging large-scale reinforcement learning and external verifiers. 
A key strategy is to train on extended sequences that incorporate chains-of-thought (CoT) or reasoning traces \citep{wei2023chainofthoughtpromptingelicitsreasoning}. 
In addition to recent advancements in post-training, emerging research explores how reasoning mechanisms might be integrated into pretraining itself \citep{geiping2025scalingtesttimecomputelatent, tack2025llmpretrainingcontinuousconcepts, ruan2025reasoninglearnlatentthoughts, dong2025reinforcementpretraining,hatamizadeh2025rlpreinforcementpretrainingobjective}. 
Rather than treating reasoning as a byproduct of scale, these approaches suggest it can instead be structured through targeted development.
A deeper understanding of how reasoning arises during training could enable us to induce it more deliberately.

\subsection{Background on interpretability and Chain-of-Thought (CoT) reasoning}
To better understand reasoning in LLMs, a plethora of recent work have turned to mechanistic explanations. 
\citet{lindsey2025biology} analyze model circuitry to trace internal reasoning patterns, though their findings are limited to carefully controlled settings.
\citet{elhage2022toymodelssuperposition}, and \citet{wang2022interpretabilitywildcircuitindirect} identify components like induction heads and algorithmic circuits, offering insights into how models support multi-step reasoning. 

However, other recent findings suggest that CoT traces may align poorly with actual inference dynamics. 
While CoT prompting has improved performance on complex reasoning tasks \citep{wei2023chainofthoughtpromptingelicitsreasoning, kojima2023largelanguagemodelszeroshot}, several studies demonstrate that its explanations can be unfaithful to the model’s true computational process \citep{turpin2023languagemodelsdontsay, Wei_Jie_2024}. Techniques such as self-consistency sampling \citep{wang2023selfconsistencyimproveschainthought}, contrastive CoT \citep{chia2023contrastivechainofthoughtprompting}, and faithful reasoning frameworks \citep{lyu2023faithfulchainofthoughtreasoning} attempt to bridge this gap. Nonetheless, the central challenge remains: how to reliably relate surface-level reasoning traces to the internal mechanisms that actually drive model behavior.

These challenges expose a key gap in reasoning datasets. Natural language CoT resources are large but unverifiable due to linguistic ambiguity. Formal corpora such as theorem-proving datasets (e.g., Lean \citep{Yang2023LeanDojoTP}, Metamath \citep{yu2024metamathbootstrapmathematicalquestions}) offer symbolic guarantees yet lack detailed annotations or remain too small for modern pretraining. Mathematical datasets like MATH \citep{hendrycksmath2021} and GSM8K \citep{cobbe2021gsm8k} provide intermediate scale but verify only final answers, not reasoning steps. No existing dataset combines programmatic step-level verification with pretraining-scale size. \textsc{FOL-Traces} fills this gap, as shown in Table~\ref{tab:dataset_comparison}.

\begin{table}[h!]
\centering
\scriptsize
\begin{tabular}{lccc}
\toprule
\textbf{Dataset} & \textbf{Scale} & \textbf{Verified Steps} & \textbf{Domain} \\
\midrule
MATH & 2.93K & \xmark\ (answers only) & Math \\
GSM8K & 8K & \xmark\ (answers only) & Math \\
Lean & <100K & \cmark & Formal proofs \\
FOL-TRACES & 7.4M & \cmark & First-order logic \\
\bottomrule
\end{tabular}
\caption{\textsc{FOL-Traces} is the only dataset combining pretraining scale with step-level verification. Scale refers to the number of examples.}
\label{tab:dataset_comparison}
\end{table}


\subsection{Formal Reasoning Presents an Opportunity}
Formal reasoning provides a clear lens for examining how language models reason. 
Unlike natural language reasoning---which is often ambiguous and hard to verify---formal logic offers interpretable, verifiable structure, yet purely symbolic tasks like arithmetic can be too limited and detached from language. 
First-order logic (FOL) strikes a useful balance: it mirrors natural-language structure while remaining algorithmically grounded and semi-decidable \citep{boolos2007computability}---valid formulas can be confirmed by an algorithm, though invalid ones cannot always be disproved. 
Leveraging this property, we introduce \textsc{FOL-Traces}, the first large-scale dataset with programmatically verified reasoning traces (7.4M examples, 3.5B tokens). 
Unlike unverifiable natural language CoT or theorem-proving corpora---which are too small for pretraining---\textsc{FOL-Traces} uniquely combines symbolic guarantees with scale, enabling systematic analysis of how models acquire, represent, and generate logical reasoning with step-by-step verification.

\paragraph{Our Contributions.} Building on this motivation, our contributions are threefold:
\begin{enumerate}
    \item We introduce \textsc{FOL-Traces}, the first large-scale dataset with programmatically verified reasoning traces (7.4M examples, 3.5B tokens)---addressing the critical gap between unverifiable natural language CoT and small-scale symbolic datasets.
    \item We propose two diagnostic tasks---\textit{masked operation prediction} and \textit{step completion}---that probe models’ syntactic awareness and process-level reasoning fidelity.  
    \item We benchmark both pretrained and reasoning-oriented LLMs, revealing systematic differences in how models acquire and internalize logical structure.
\end{enumerate}

\paragraph{Scope.} \textsc{FOL-Traces} is intended as a diagnostic complement to natural language reasoning benchmarks rather than a replacement for them.
By operating with programmatically verified FOL steps, it isolates core logical reasoning inference behavior while deliberately abstracting away linguistic ambiguity. FOL \textit{does} represent a subset of natural language, but our dataset surgically, intentionally focuses on verifiable structured inference and does not aim to capture the full richness of open-ended natural language reasoning.

\section{\textsc{FOL-Traces} dataset}

\begin{figure}[!h]
  \centering
  \includegraphics[width=0.98\linewidth]{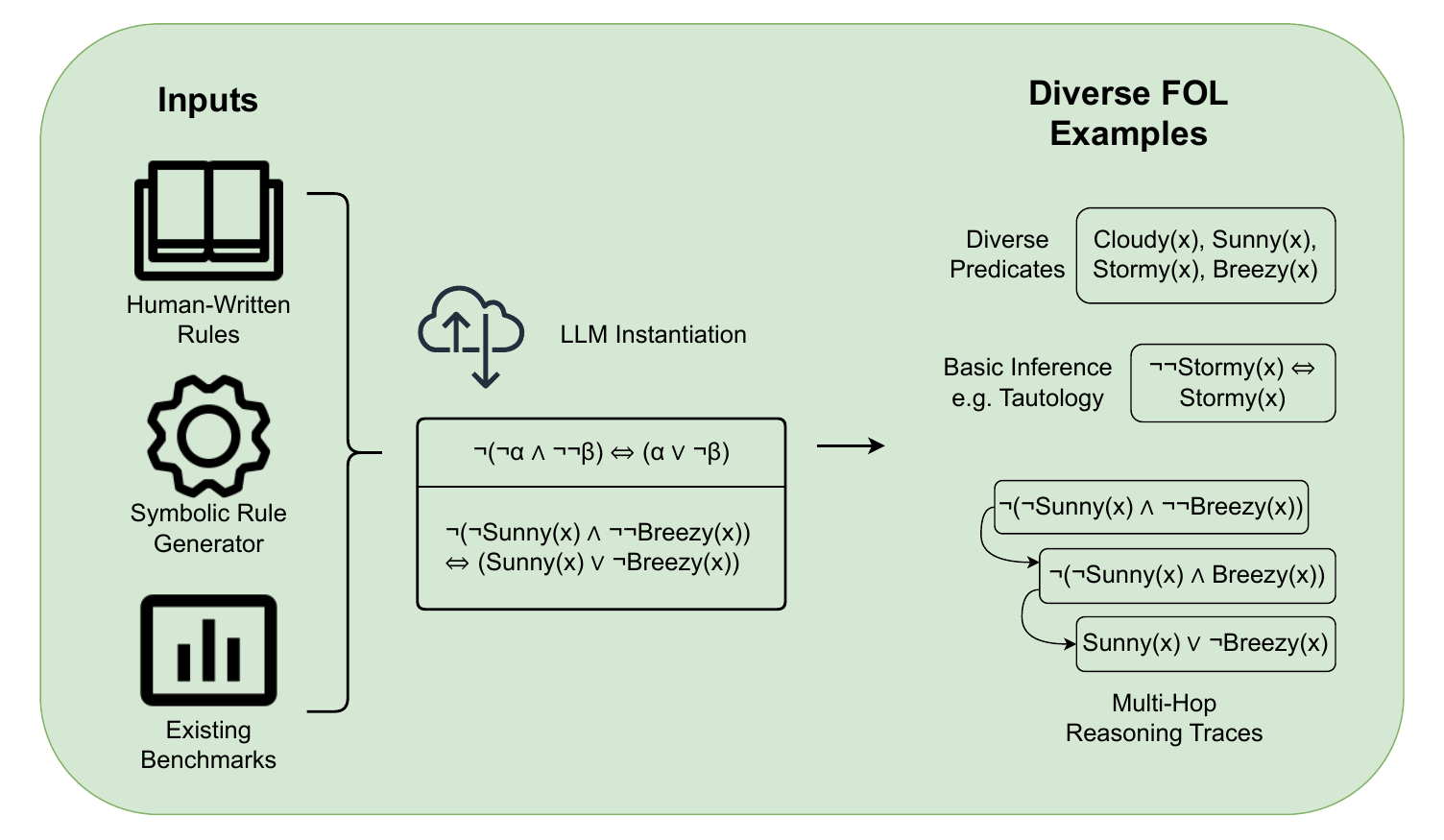}
  \caption{FOL Dataset Generation Overview.}
  \label{fig:visual-abstract}
\end{figure}

Most reasoning datasets face a trade-off: natural language CoT is unverifiable, while symbolic logic corpora are too small for large-scale training. \textsc{FOL-Traces} combines programmatic verification with pretraining-scale size, enabling systematic study of logical reasoning. The generation pipeline shown in Figure~\ref{fig:visual-abstract} (1) generates symbolic FOL formulas and (2) instantiates them with natural language predicates and variables via LLMs.

We leverage existing benchmarks such as LogicBench~\cite{parmar2024logicbenchsystematicevaluationlogical} and FOLIO~\cite{han2024folionaturallanguagereasoning} as sources of human-annotated logical forms and prompts. 
In addition to the collected inference rules from existing datasets, we programmatically generate 1.5 million unique rules and corresponding step-by-step simplification traces.
Then, we employ LLMs to instantiate them, resulting in a diverse and extensive collection of FOL expressions. 
We describe the programmatic unique rule generation process in \S\ref{sec:generation-with-sympy}. Then, we detail the various properties and the complexity dynamics of the rule sets and resulting datasets in \S\ref{sec:dataset-distribution}.

\subsection{Dataset Structure and Annotations} 

The dataset contains pretraining-scale training splits, as well as comprehensively labeled validation and test splits. 
Each example is grounded in an underlying first-order symbolic expression annotated with a unique rule ID. 
Subsequently, each rule is annotated in detail with the following fields. 

\textbf{Rule (\texttt{rule}).} This field contains the first-order logic expression template to be instantiated. Each rule consists of at least one first-order logic formula. For curated rules, we use basic inference rules from first-order logic. For randomly generated rules, we programmatically generate a formula and then simplify it step by step to create a chain of first-order logic expressions that together form a rule.

\textbf{Expressions (\texttt{exprs}).}
For randomly generated rules, each formula is simplified using basic first-order logic inference rules. Each simplification step is recorded as an expression and added to the list in this field.

\textbf{Complexity by step (\texttt{complexity\_by\_step}).} 
The circuit complexity corresponding to each expression in \texttt{exprs}.
Here, we define the circuit complexity of a Boolean or first-order logical formula \( \varphi \) with

\begin{itemize}
    \item \( \text{Atoms}(\varphi) \): the set of atomic propositions in \( \varphi \) (i.e., statements with Boolean values \textsc{True}/\textsc{False}),
    \item \( \text{Sub}(\varphi) \): the set of \emph{immediate subformulas} of~$\varphi$, 
    i.e., those that occur exactly one level below~$\varphi$ in its syntactic tree. 
    For example, if $\varphi = (\alpha \land \neg \beta) \lor \gamma$, then 
    $\text{Sub}(\varphi) = \{\alpha \land \neg \beta, \gamma\}$.
    \item Each logical connective (e.g., \( \neg, \land, \lor \)) corresponds to a logic gate in a Boolean circuit.
\end{itemize}

Then, we define the circuit complexity as the circuit size \( C(\varphi) \) needed to compute \( \varphi \). Formally, we define as:

\begin{equation}
C(\varphi) =
\begin{cases}
1, & \text{if } \varphi \text{ is atomic} \\
1 + \sum\limits_{i=1}^{k} C(\varphi_i), & \text{if } \varphi = f(\varphi_1, \ldots, \varphi_k)
\end{cases}
\label{eq:circuit-complexity}
\end{equation}
\noindent
where \( f \) is a logical operator (e.g., \( \neg, \land, \lor \)), and \( \varphi_1, \ldots, \varphi_k \in \text{Sub}(\varphi)\).
Generally, the circuit complexities decrease for subsequent expressions in \texttt{exprs}, as they simplify.

\textbf{\mbox{Elimination complexity} (\path{elimination_complexity}).}
Each expression in \texttt{exprs} is a simplification of the one before it. 
The elimination complexity at index~$i$ is the program complexity of simplifying \texttt{exprs[$i$]} into \texttt{exprs[$i{+}1$]}.  \looseness=-1

\textbf{Program complexity (\texttt{program\_complexity}).}  
This represents the total program complexity involved in randomly generating a first-order logic expression and subsequently, recursively simplifying it. In some cases, the expression is fully simplified. However, if the simplification requires too large a recursive jump, the program may stop prematurely. A detailed description of the full algorithm is provided in \S\ref{sec:generation-with-sympy}.

\textbf{Original depth (\texttt{original\_depth}).}  
The original depth is the depth of the initial expression, i.e., \texttt{exprs[0]}. 

Finally, Table~\ref{tab:dataset_stats} presents summary statistics for the dataset splits. Compared to other datasets, our reasoning CoT traces are substantially larger by a wide margin. Moreover, all of these existing datasets lack detailed annotations for assessing correctness or for quantifying different types of reasoning failures.

\begin{table}[h!]
\centering
\scriptsize
\begin{tabular}{l l c c}
\toprule
\textbf{Dataset} & \textbf{Split} & \textbf{\# Examples} & \textbf{\# Tokens} \\
\midrule
\multirow{4}{*}{\textbf{\textsc{FOL-Traces}}} 
 & Train & 7.40M & 3.51B \\
 & Dev & 5K & 3.05M \\
 & Test & 8.64K & 5.25M \\
 & Curated Train & 8.80M & 786.27M \\
\bottomrule
\end{tabular}
\caption{Number of examples and tokens in each dataset split. \textsc{FOL-Traces} splits are generated from synthetic rules produced by the SymPy program. The curated training set was generated from human-annotated benchmarks augmented with LLM prompting. }
\label{tab:dataset_stats}
\end{table}

\subsection{Dataset Generation}

We constructed the dataset through a hybrid approach combining human-curated rules, automated symbolic rule generation, and LLM prompting. The detailed breakdown of the inference rules as well as human curated rules are shown in \S\ref{sec:fol-categories}, \S\ref{sec:fol-complex-categories}, \S\ref{sec:fol-eliminations-categories}, and \S\ref{sec:table_training_data}. Our process began with the collection of existing benchmarks for logical reasoning, which provided both direct training material and a basis for generating generalized logical proofs via LLMs.

To expand the diversity of logical forms, we developed templates representing canonical inference rules in first-order logic, including De Morgan’s laws, tautologies, and implications. These templates were instantiated with meaningful predicates using LLMs. For example, the general form
\begin{equation}
\neg(\alpha \lor \beta) \rightarrow \neg \alpha \land \neg \beta
\end{equation}
was instantiated as
\begin{multline}
\neg(\text{Sunny}(x) \lor \text{Breezy}(x)) \rightarrow \\
\neg \text{Sunny}(x) \land \neg \text{Breezy}(x)
\end{multline}

where $\text{Sunny}(x)$ and $\text{Breezy}(x)$ are predicates describing environmental properties. Such instantiations were generated by prompting GPT-4o with the abstract logical form and a set of domain-relevant predicate candidates.
Additionally, we used symbolic computation tools such as SymPy to synthesize first-order logic expressions, ensuring both syntactic correctness and logical validity. This multi-stage generation process enabled the creation of a high-quality, diverse dataset designed to evaluate and train models on robust logical reasoning tasks.

\subsubsection{Generate rules with SymPy}
\label{sec:generation-with-sympy}
To generate verifiably correct first-order logic expressions, we utilize SymPy. The generation process follows the algorithm outlined in Algorithm~\ref{alg:random_formula}, which recursively constructs logical formulas based on a random selection of operators and symbols. 

\begin{algorithm}[t]
\caption{Generate a Random Formula}
\label{alg:random_formula}
\footnotesize
\algrenewcommand\algorithmicindent{0.8em}
\begin{algorithmic}[1]
\Function{RandomFormula}{$\Sigma$, $d$, $m$, $n$}
  \If{$d=0$} \State \Return random choice from $\Sigma$, $n{+}1$ \EndIf
  \State $op \gets \text{rand}\{\land,\lor,\lnot,\rightarrow\}$
  \If{$op=\lnot$}
    \State $sub,n \gets \Call{RandomFormula}{\Sigma, d{-}1, m, n}$
    \State \Return $\lnot sub$, $n{+}1$
  \Else
    \State $L,n \gets \Call{RandomFormula}{\Sigma, d{-}1, m, n}$
    \State $R,n \gets \Call{RandomFormula}{\Sigma, d{-}1, m, n}$
    \State \Return $op(L,R)$, $n{+}1$
  \EndIf
\EndFunction
\end{algorithmic}
\end{algorithm}

The depth and complexity of the formula are controlled by the parameters provided. Once a formula is generated, a high-level simplification procedure is applied (described in Algorithm~\ref{alg:simplify_formula}) to reduce the formula to its simplest form by applying standard logical inference rules such as De Morgan’s laws and eliminating tautologies. The full pseudocode for these processes is provided in \S\ref{sec:full-algo}. 

\begin{algorithm}[t]
\caption{High-Level Subexpression Simplification}
\label{alg:simplify_formula}
\footnotesize
\algrenewcommand\algorithmicindent{0.8em}
\begin{algorithmic}[1]
\Function{Simplify}{$F, d$}
  \If{depth($F$) $<$ threshold}
    \State simplify $F$ directly
  \Else
    \State recursively simplify sub-parts
  \EndIf
  \If{$F \in \{\lnot,\land,\lor,\rightarrow\}$}
    \State apply logical rules (e.g., De Morgan, tautology)
  \EndIf
  \State \Return simplified $F$
\EndFunction
\end{algorithmic}
\end{algorithm}

Finally, we generate chains of reasoning traces by sequentially linking the simplification steps. This process is akin to CoT reasoning, where each intermediate reasoning step builds upon the previous one, forming a structured sequence that leads to the final conclusion. However, unlike traditional reasoning traces, our approach ensures that each step is programmatically verified to be correct, leveraging SymPy's symbolic computation capabilities. By chaining these simplifications, we capture a stepwise progression of logically sound transformations, which can be useful for tasks that require a detailed, interpretable sequence of logical reasoning.

\section{Dataset Statistics}
\label{sec:dataset-distribution}

\begin{figure}[h]
  \centering
  \includegraphics[width=\columnwidth]{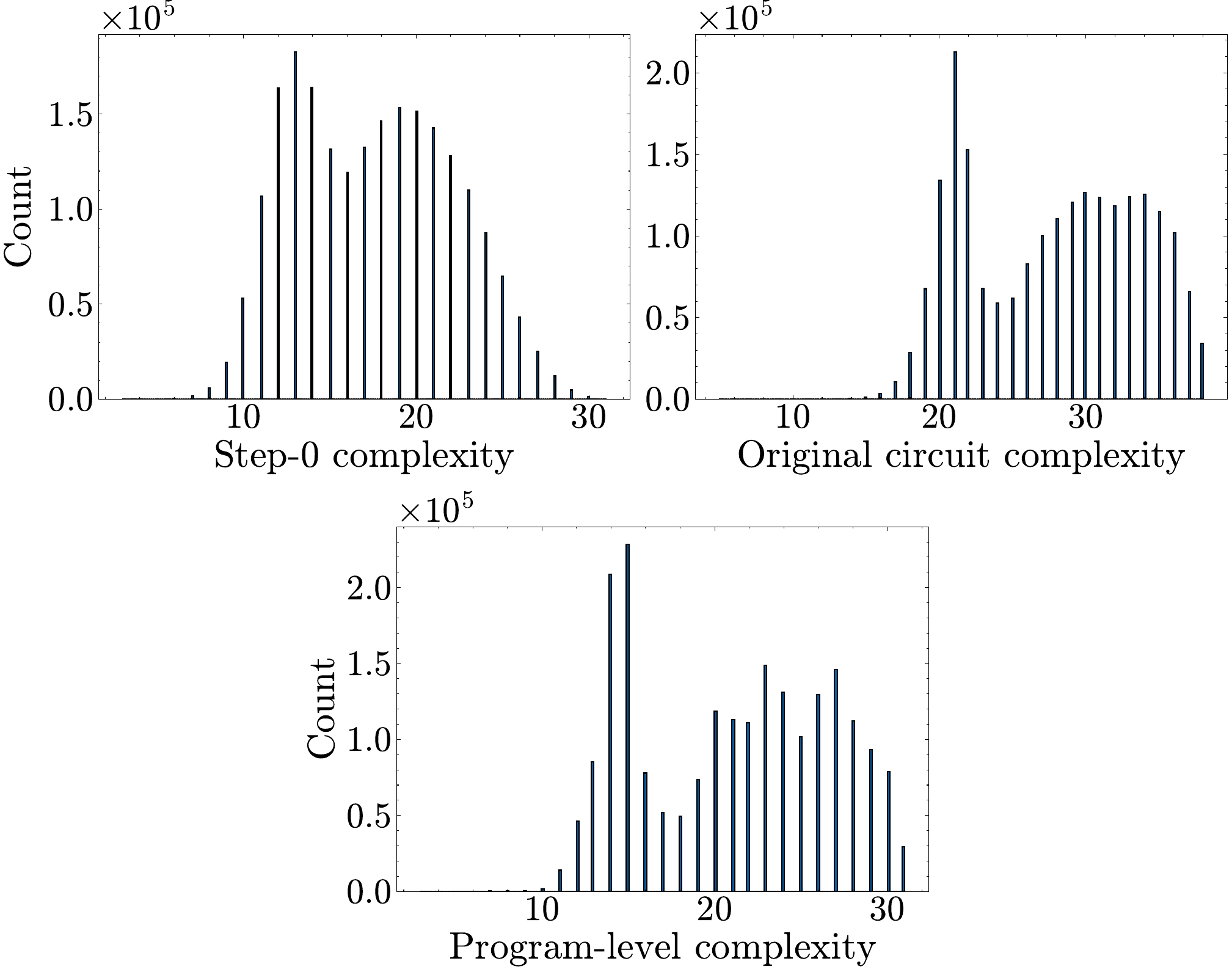}
  \caption{Rule‐level complexity distributions.}
  \label{fig:rule_complexity_dists}
\end{figure}

We begin by examining the distribution of rules in our dataset, which were generated to provide a diverse set of templates for constructing complexity-annotated examples used in pretraining. A summary of this distribution is presented in Figure~\ref{fig:rule_complexity_dists}. 

Figure~\ref{fig:rule_complexity_dists} illustrates the distribution of the circuit complexity \( C(\varphi) \), as defined in Equation~\ref{eq:circuit-complexity}, at step 0. The distribution appears to be a bimodal normal distribution with the mean around 13 and 19. Then, the original circuit complexity in Figure~\ref{fig:rule_complexity_dists} describes \( C(\varphi) + \text{maximum depth of circuit} + \text{number of unique variables}\). Notably, the distribution appears tri-modal, with a concentration of low-complexity rules around 20, a higher-complexity group near 30, and an overall mean complexity around 35. Then, as the rule simplifies, the total program complexity to generate the rule chains propagates as a trimodal distribution, with mean at 15, 23, and 27.

\begin{figure}[h]
  \centering
  \includegraphics[width=\linewidth]{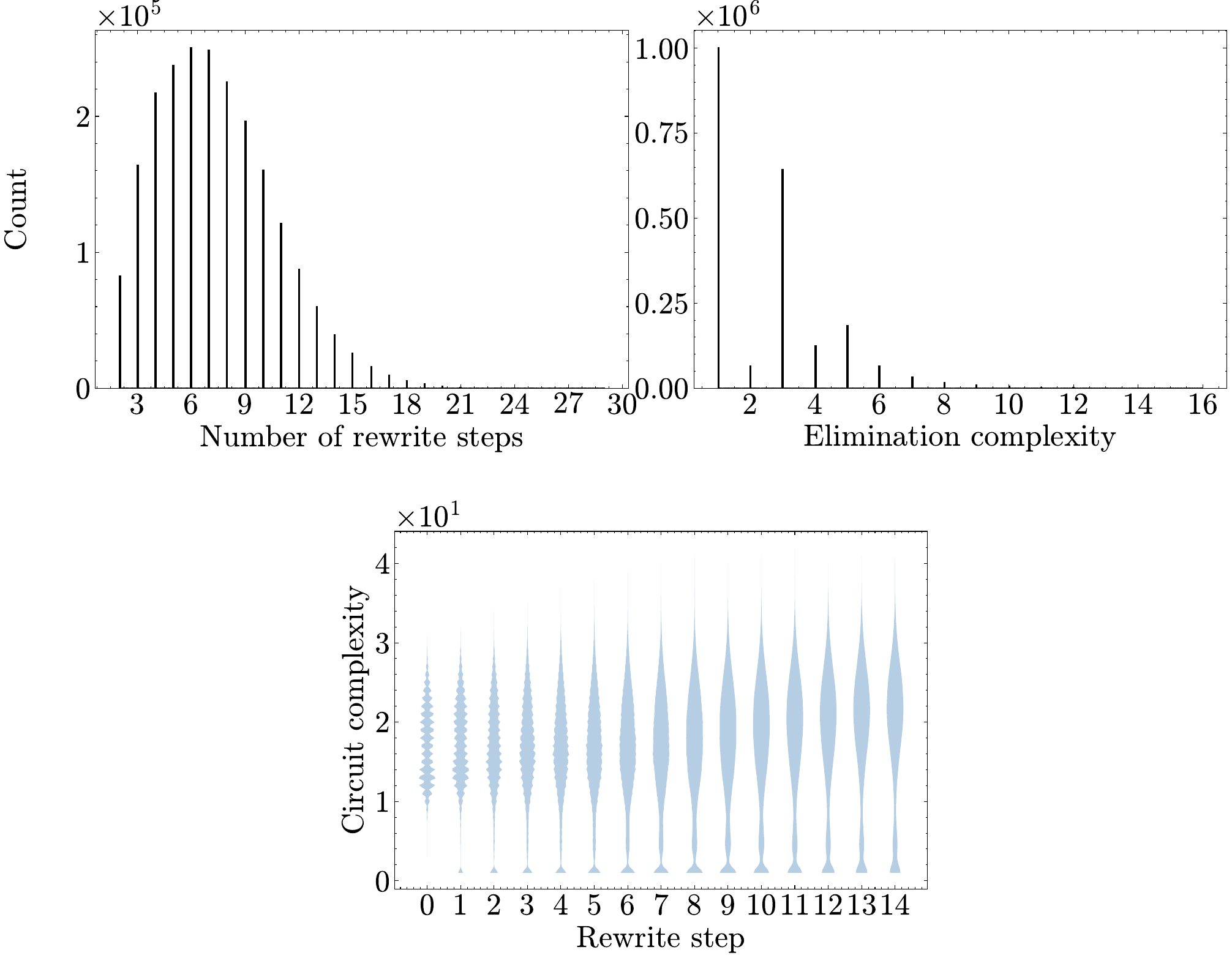}
  \caption{Rule‐level dynamics. Top left: distribution of rewrite steps per rule. Top right: elimination complexity at step 0. Bottom: circuit complexity across rewrite steps.   Together, these illustrate how complexity evolves and simplifies during the rewrite process.}

  \label{fig:rule_dynamics_combined}
\end{figure}

Next, we examine how rewrite rules evolve throughout the simplification process, as illustrated in Figure~\ref{fig:rule_dynamics_combined}. The top left panel shows that the number of rewrite steps per rule follows an approximately normal distribution with a long tail, indicating that most rules simplify quickly while a few require many steps. The top right panel depicts elimination complexity at step~0, which is typically low but occasionally high, reflecting rare difficult simplifications. The bottom panel shows that circuit complexity generally increases with rewrite depth, suggesting that longer rewrite sequences correspond to more intricate transformations.

\begin{figure}[t] 
  \centering
  \includegraphics[width=\columnwidth]{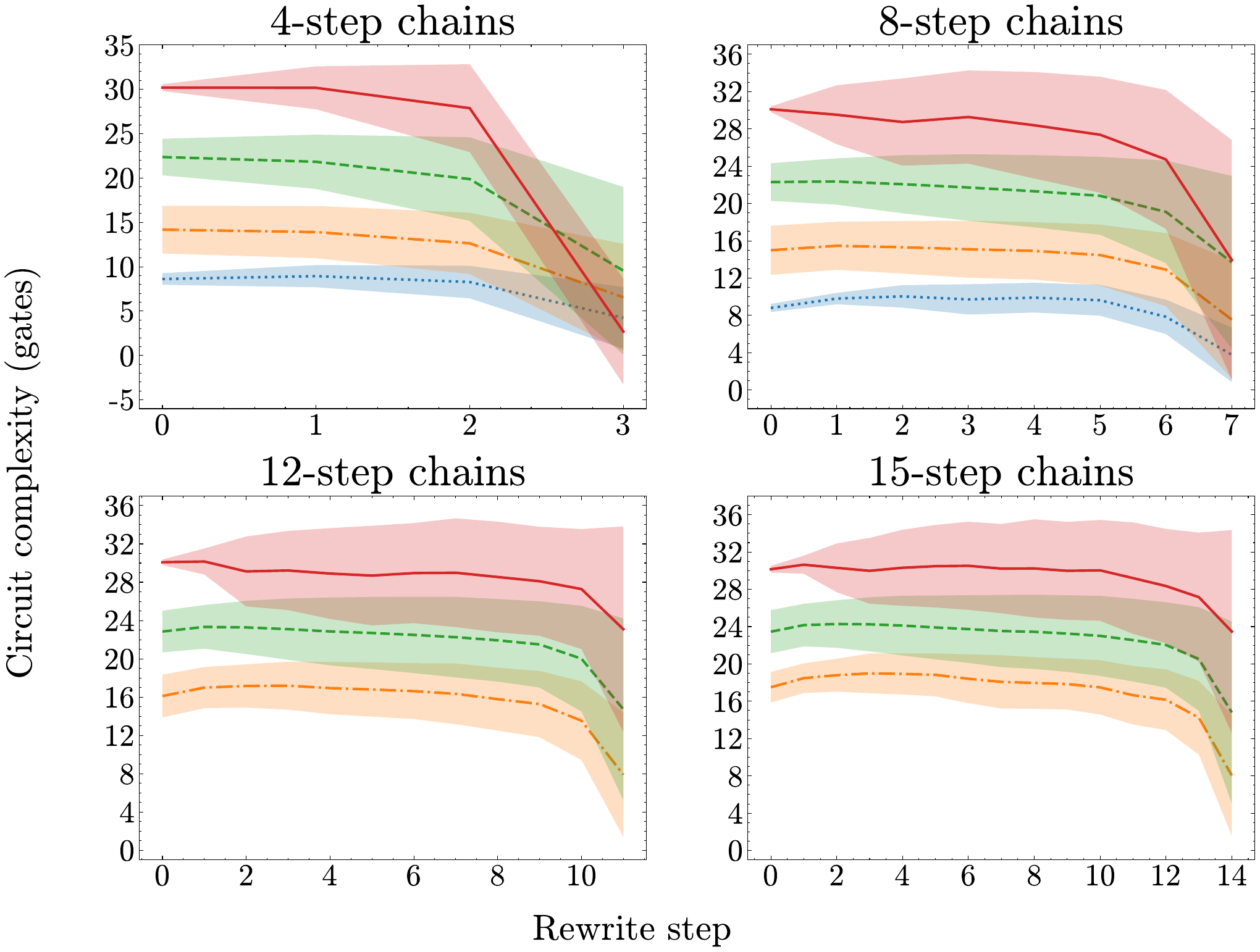}
  \caption{Circuit‐complexity evolution by chain length.
    For each fixed‐length bucket, we group traces by their starting gate complexity:  
    \emph{blue solid} = $[0-9]$,  
    \emph{orange dashed} = $[10-19]$,  
    \emph{green dash‐dotted} = $[20-29]$,  
    \emph{red dotted} = $\geq 30$ gates.  
    Shaded bands show $\pm 1\sigma$ around the mean.}
  \label{fig:example}
\end{figure}

Furthermore, we analyze how reasoning trace complexity changes across elimination steps (Figure~\ref{fig:example}). Complexity generally decreases, with a sharp drop near the end, and final expressions show greater variation than initial ones. Figure~\ref{fig:train2_tokens} summarizes token counts per training example.

\section{Experiments} 
\label{sec:exp}

We conduct two stages of experiments to assess both the stability of our training setup and the reasoning competence it supports. 
First, we perform pretraining sanity checks to verify that the \textsc{FOL-Traces} corpus enables stable optimization and coherent FOL representations in a small LM. 
We then test this initial pretraining on a one-step \textsc{True}\,/\,\textsc{False} evaluation as a minimal reasoning benchmark, as proxies for more complicated FOL tasks we introduce with current reasoning-oriented models.
Second, we evaluate reasoning-oriented models on two structured diagnostic tasks---(1) masked operation prediction and (2) step completions---that probe syntactic awareness and process fidelity, allowing systematic comparison across model families and scales.

\subsection{Pretraining Sanity Checks}

As a preliminary step, we pretrain \citet{Karpathy2022}'s implementation of the 125M-parameter GPT-2-small architecture (12 layers, 12 heads per layer) from scratch on the \textsc{FOL-Traces} corpus to validate that the dataset supports stable optimization and coherent FOL representations prior to downstream reasoning evaluations. The model uses an embedding size of 768, a block size of 1024 tokens, and a micro-batch size of 12 with 40 gradient accumulation steps to simulate a larger effective batch. Training employs the AdamW optimizer with a learning rate of $1\times10^{-4}$, weight decay of 0.1, gradient clipping at 1.0, and no dropout. The learning rate follows a warm-up over 400 iterations and decays to a minimum of $1\times10^{-5}$ across 20{,}000 iterations. Training was conducted on two NVIDIA RTX A6000 GPUs for approximately 40~hours. 

We conduct two complementary studies---one-step \textsc{True}\,/\,\textsc{False} evaluation as a basic reasoning task and representation-level probing---to assess how the 125M-parameter GPT-2 variant internalizes FOL.

\paragraph{One-Step \textsc{True}\,/\,\textsc{False} Evaluation}
\label{sec:1step-discussion}

As a quick diagnostic, we analyze one-step reasoning with definitive truth values, focusing on the simplest strictly evaluable forms of reasoning.
For each formula–interpretation pair \((\varphi,\mathcal I)\), we prompt the model with ``$\texttt{\string<bos\string>}\,\varphi \Leftrightarrow \texttt{\_\_}$'' and count a prediction correct when the next token matches the gold \textsc{True} or \textsc{False}. Although the full 50{,}304 vocabulary is available, the model rapidly collapses its output distribution onto a binary subspace, indicating early acquisition of the truth-evaluation task.

To analyze difficulty, we partition the evaluation set by original complexity into three buckets: low, medium, and high. Figure~\ref{fig:top1_accuracy_complexity} traces Top-1 accuracy across training tokens, and Table~\ref{tab:learning_dynamics} reports the tokens required to reach 70\% accuracy on the unseen split as well as the final accuracies. Syntax is learned rapidly: after $5\times10^{8}$ tokens (5\% of training), accuracy improves from 50\% to 60\%, indicating that the model collapses its output space to a binary decision. 


\begin{table}[h]
\centering
\small
\begin{tabular}{lccc}
\toprule
Bucket & Tokens to 70\% & Final (seen) & Final (unseen)\\
\midrule
0–21 & 3.2 B & 0.76 & 0.76\\
22–32 & 1.1 B & 0.79 & 0.87\\
$\ge$ 33 & 1.1 B & 0.89 & 0.92\\
\bottomrule
\end{tabular}
\caption{Learning dynamics on one-step \textsc{True}/\textsc{False}. `Tokens to 70\%' refers to the \textsc{unseen} split.}
\label{tab:learning_dynamics}
\end{table}

\paragraph{Probing Geometric Sensitivity to Syntactic Validity}
LLMs are typically evaluated on task accuracy, which overlooks whether their hidden states encode syntactic well-formedness. 
We recast our \textsc{True}\,/\,\textsc{False} corpus into a binary validity benchmark, creating minimal corruptions by deleting one token from fifty \textsc{True} formulas to disrupt global coherence.
We compare our pretrained NanoGPT with BERT-base, and compute the cosine similarities between the predicted embeddings and the valid labels. 
We quantify the Spearman's $\rho$ bootstrapped over 1{,}000 samples. 
NanoGPT’s embeddings showed a much stronger alignment with syntactic validity ($\rho \approx 0.68$) than BERT’s ($\rho \approx 0.33$), likely because GPT’s autoregressive training enforces a kind of ``syntactic checksum'' that quickly detects missing or malformed tokens, whereas BERT’s bidirectional masking spreads error signals across the sequence, weakening its sensitivity to such structural breaks. Overall, this compact FOL corpus and embedding-level RSA reveal that a small language model can encode syntactic structure geometrically when trained with \textsc{FOL-Traces}.

\begin{table}[h]
\centering
\resizebox{\columnwidth}{!}{%
\begin{tabular}{lccc}
\toprule
Model & Representation & $\rho$ & 95\% CI \\
\midrule
NanoGPT (20\,k) & final BOS stream & \textbf{0.682} & [0.666, 0.695] \\
BERT-base       & final \texttt{[CLS]} & 0.327 & [0.308, 0.345] \\
\bottomrule
\end{tabular}}
\caption{RSA between embedding similarities and validity labels.}
\label{tab:rsa_validity}
\end{table}

\subsection{Structured Reasoning Diagnostic Tasks}
We introduce and perform two diagnostic tasks---\textit{masked operation prediction} and \textit{step completion}---that directly probe syntactic awareness and process fidelity. 
These tasks are designed to isolate core reasoning behaviors: the masked operation prediction task targets atomic-level agreement in first-order logic syntax, 
while the step completion task evaluates the model's ability to extend reasoning chains consistent with first-order logic. 
Both tasks are automatically evaluable via programmatic verifiers, enabling scalable and objective assessment.
Experiments are conducted using five reasoning models from two families, 
Phi-3~\citep{abdin2024phi3technicalreporthighly} and Qwen-2.5~\citep{bai2023qwentechnicalreport}, 
covering a range of model sizes.

\subsubsection{Task 1: Masked Operation Prediction}
\begin{figure}[!h]
  \centering
  \includegraphics[width=0.5\linewidth]{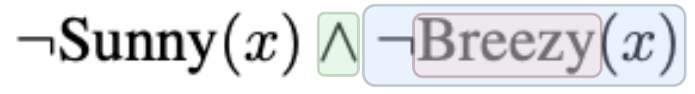}
  \caption{Masked regions for Task 1 for an example FOL phrase, showing \textit{component} (blue), \textit{operator} (green), and \textit{predicate} (red) areas used in the respective subtasks.}
  \label{fig:task1-subtask-masks}
\end{figure}

For a targeted evaluation of distinct aspects of compositional reasoning, we isolate specific dimensions of the model’s compositional understanding through targeted input masking. As illustrated in Figure~\ref{fig:task1-subtask-masks}, Task 1 employs three masking schemes to separate reasoning components: component masks (blue) obscure subexpressions or entities, operator masks (green) conceal logical or relational connectors, and predicate masks (red) hide function or action tokens. This setup enables controlled analysis of the model’s internal representations, revealing how each subtask contributes to the overall reasoning process. The masked operation prediction tasks probe distinct facets of compositional reasoning: component accuracy measures the model’s ability to correctly identify and process complete subexpressions within logical expressions; operator accuracy evaluates whether the model correctly applies logical or relational operators between components; and predicate accuracy assesses correctness in identifying function names. Together, these metrics provide a holistic view of the model’s systematic reasoning capabilities.

Across all evaluated models, performance varies considerably by both model family and size. Within Phi-3 models, accuracy remains relatively low across all metrics, with \textit{phi-3-medium-128k} performing particularly poorly and \textit{phi-3.5-mini} showing only modest gains. In contrast, the Qwen-2.5 series demonstrates a clear scaling trend: larger models consistently outperform smaller ones, culminating in \textit{qwen2.5-32b}, which achieves the highest scores in every category: 75\% component accuracy, 73\% operator accuracy, and 80\% predicate accuracy. Despite these improvements, even the strongest models show limited robustness and generalization across compositional and operational reasoning tasks. Notably, models explicitly trained for reasoning, such as \textit{phi-3.5-mini}, still underperform substantially compared to larger general-purpose models. This suggests that current reasoning-oriented training approaches alone may be insufficient for achieving strong systematic reasoning performance, highlighting the need for more integrated reasoning and compositional generalization strategies.

\begin{table}[h]
\centering
\scriptsize
\setlength{\tabcolsep}{3pt}
\renewcommand{\arraystretch}{1.1}
\begin{tabular}{lcccc}
\toprule
\textbf{Model} & \textbf{Comp. Acc} & \textbf{Oper. Acc} & \textbf{Pred. Acc} & \textbf{Overall} \\ 
\midrule
phi-3.5-mini       & 26.00\% & 32.00\% & 19.00\% & 25.67\% \\
phi-3-medium-128k  & 1.00\% & 22.00\% & 17.00\% & 13.33\% \\
\midrule
qwen2.5-7b         & 64.00\% & 25.00\% & 62.00\% & 50.33\% \\
qwen2.5-14b        & 72.00\% & 42.00\% & 75.00\% & 63.00\% \\
qwen2.5-32b        & \textbf{75.00\%} & \textbf{73.00\%} & \textbf{80.00\%} & \textbf{76.00\%} \\
\midrule
\textbf{Overall} & 47.60\% & 38.80\% & 50.60\% & 45.67\% \\
\bottomrule
\end{tabular}
\caption{Model accuracies across components, operators, and predicates prediction tasks.}
\label{tab:model-accuracy}
\end{table}

\subsubsection{Task 2: Step Completion}

For the \textit{step completion} task, we ask the models to complete the last couple steps given a FOL chain. Using SymPy as an automatic verifier, we programmatically check whether the generated reasoning steps are logically equivalent to the gold steps. Table~\ref{tab:last-two-step-completion-results} summarizes the results, showing that the Qwen2.5 models outperform Phi-3. However, even the best-performing Qwen2.5-32B model struggles to surpass 50\% accuracy, indicating that this step completion task remains highly challenging for large reasoning models.

\begin{table}[h!]
\centering
\scriptsize
\begin{tabular}{lccc}
\toprule
\textbf{Model} & \textbf{1-Step} & \textbf{2-Step} & \textbf{2-Step Chain} \\
\midrule
phi-3.5-mini        & 0.10 & 0.00 & 0.00 \\
phi-3-medium-128k   & 0.10 & 0.08 & 0.20 \\
\midrule
qwen2.5-7b          & 0.54 & 0.22 & 0.44 \\
qwen2.5-14b         & 0.60 & 0.26 & 0.48 \\
qwen2.5-32b         & \textbf{0.76} & \textbf{0.40} & \textbf{0.48} \\
\midrule
\textbf{Overall}              & 0.43 & 0.16 & 0.27 \\
\bottomrule
\end{tabular}
\caption{Model accuracies across models for last 1- and 2-step completion tasks}
\label{tab:last-two-step-completion-results}
\end{table}

\begin{figure}[h]
    \centering
    \includegraphics[width=\linewidth]{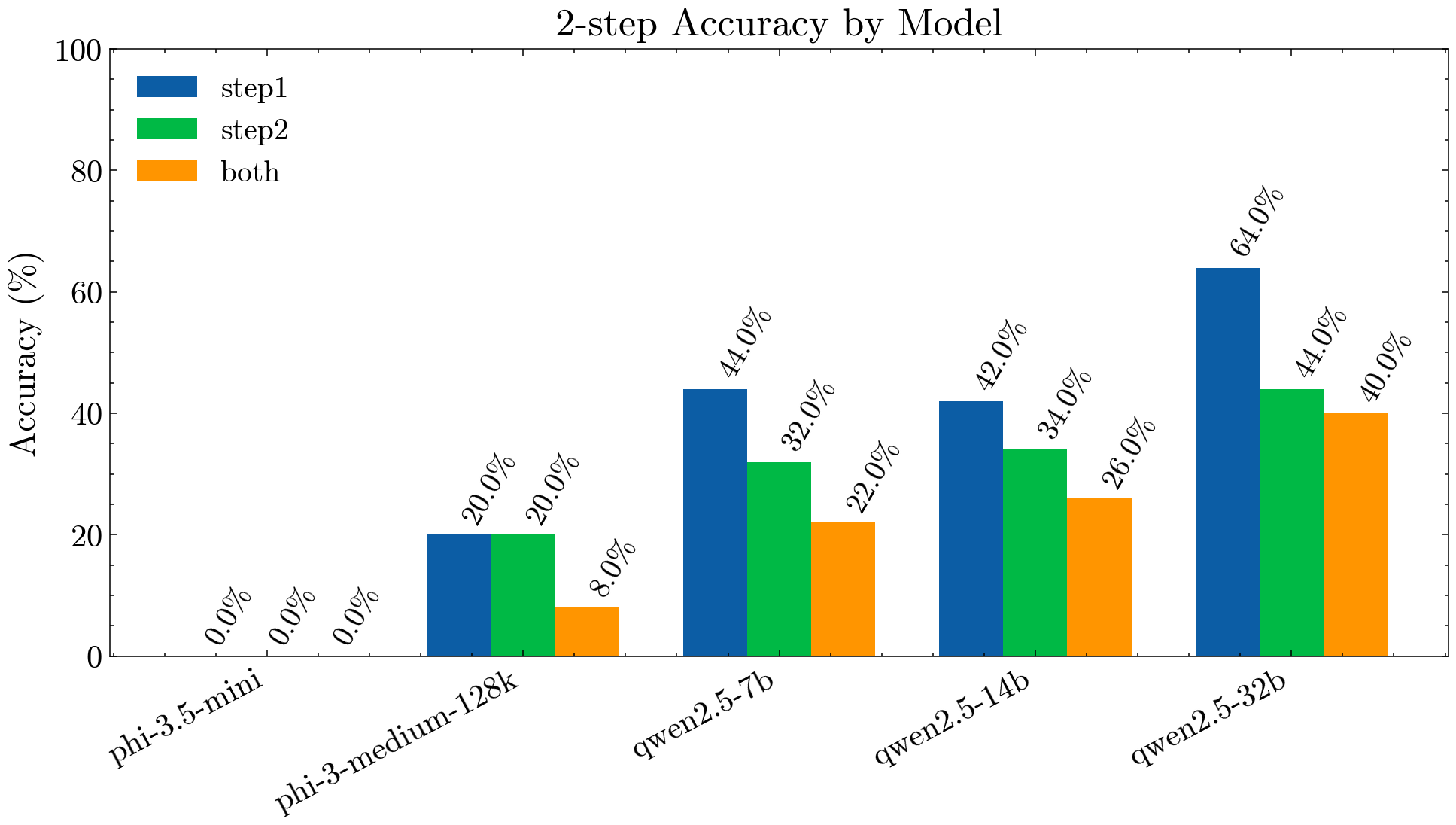}
    \caption{The last 1- and 2-step accuracies by model.}
    \label{fig:task-2-accs}
\end{figure}

A more detailed break down for 2-steps completion is shown in Figure~\ref{fig:task-2-accs}. We see that all the models perform on the last step prediction the better than the two step prediction.

\begin{figure}[t]
  \centering
  \begin{subfigure}{\columnwidth}
    \centering
    \includegraphics[width=\columnwidth]{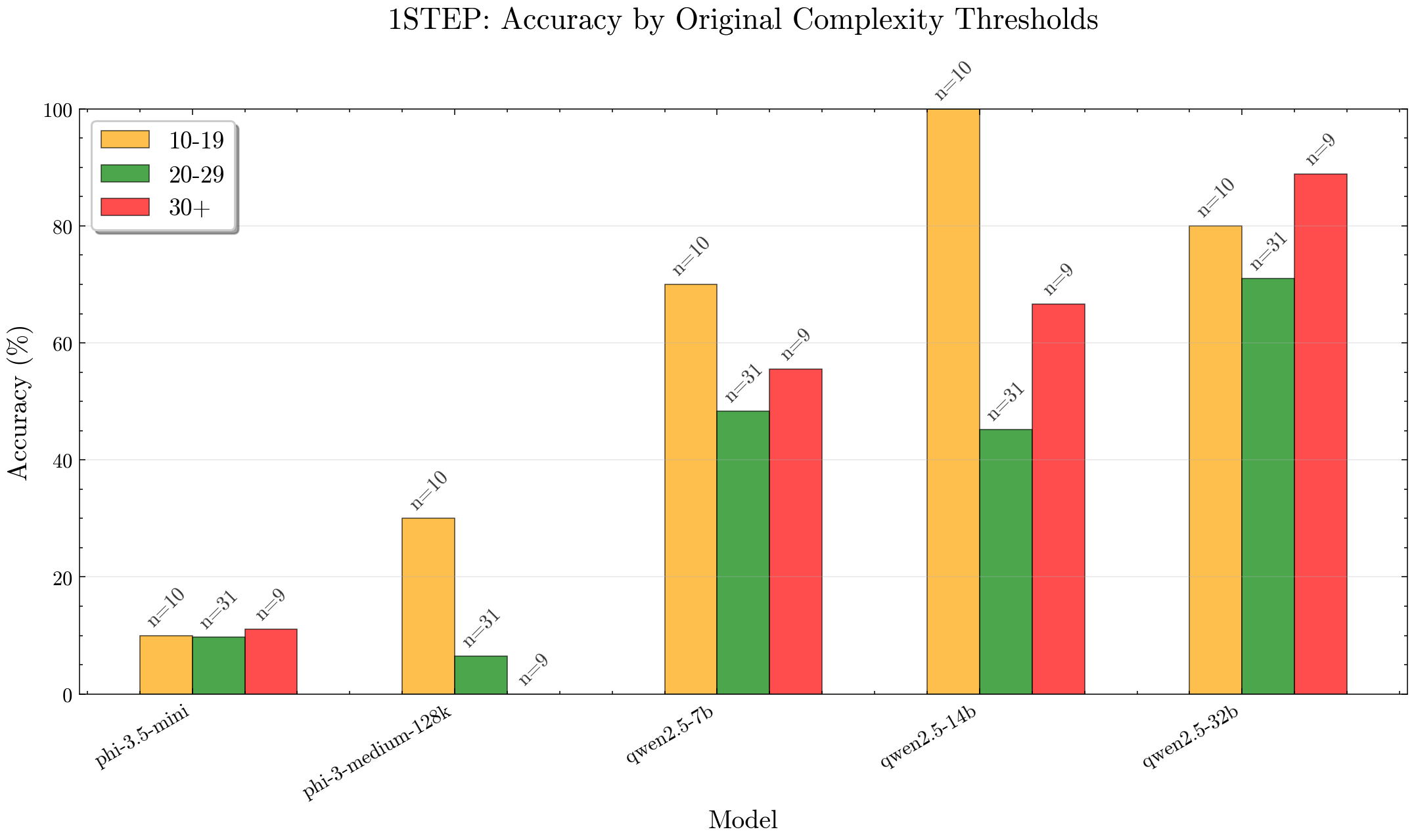}
    \caption{Last 1-step completion accuracy by original complexity thresholds.}
    \label{fig:step-1-acc-by-complexity}
  \end{subfigure}

  \vspace{0.75em}

  \begin{subfigure}{\columnwidth}
    \centering
    \includegraphics[width=\columnwidth]{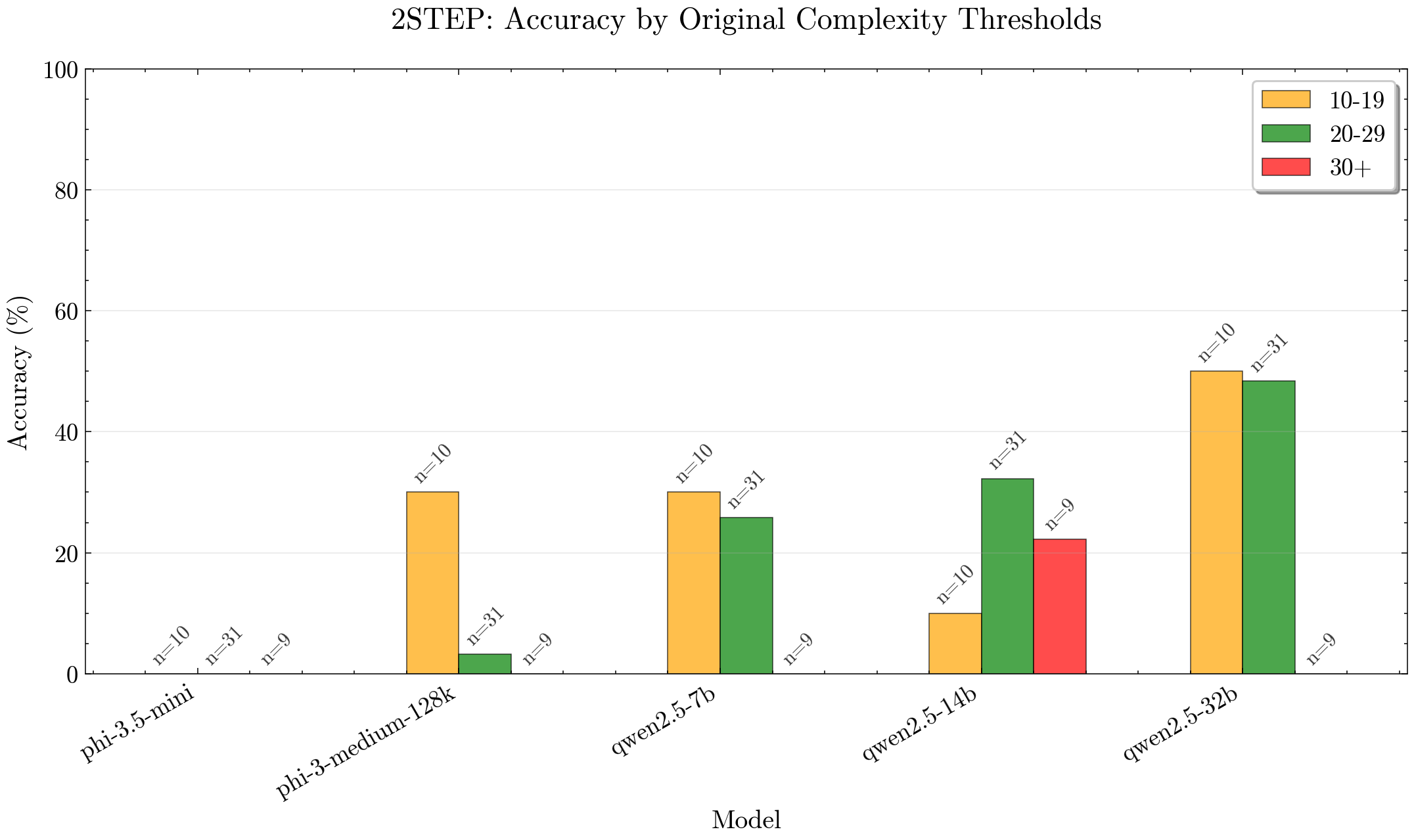}
    \caption{Last 2-step completion accuracy by original complexity thresholds.}
    \label{fig:step-2-acc-by-complexity}
  \end{subfigure}

  \caption{Accuracy across the two diagnostic tasks. Bars indicate performance over three complexity ranges (10--19, 20--29, 30+).}
  \label{fig:step-acc-by-complexity}
\end{figure}

We further stratify the results based on the original complexities of the underlying symbolic FOL rules, as shown in Figure~\ref{fig:step-acc-by-complexity}. For both the last 1- and 2-step completion tasks, LLMs achieve higher accuracy on low-complexity rules (yellow), while their performance declines markedly as rule complexity increases. As illustrated in Figure~\ref{fig:step-1-acc-by-complexity}, several reasoning LLMs are able to correctly generate low-complexity rules; however, their accuracy drops substantially for medium (green)- and high (red)-complexity examples. Similarly, Figure~\ref{fig:step-2-acc-by-complexity} shows that recent reasoning LLMs perform poorly even on relatively simple cases---with the best-performing model, Qwen2.5-32B, attaining only around 50\% accuracy for low- and medium-complexity examples. Most models fail to generate correct outputs for high-complexity cases (with the exception of Qwen2.5-14B, which reaches approximately 20\% accuracy), underscoring that this task---and derivative $n$-step completion tasks---would remain highly challenging for current LLMs.

\begin{figure}[!h]
  \centering
  \includegraphics[width=\columnwidth]{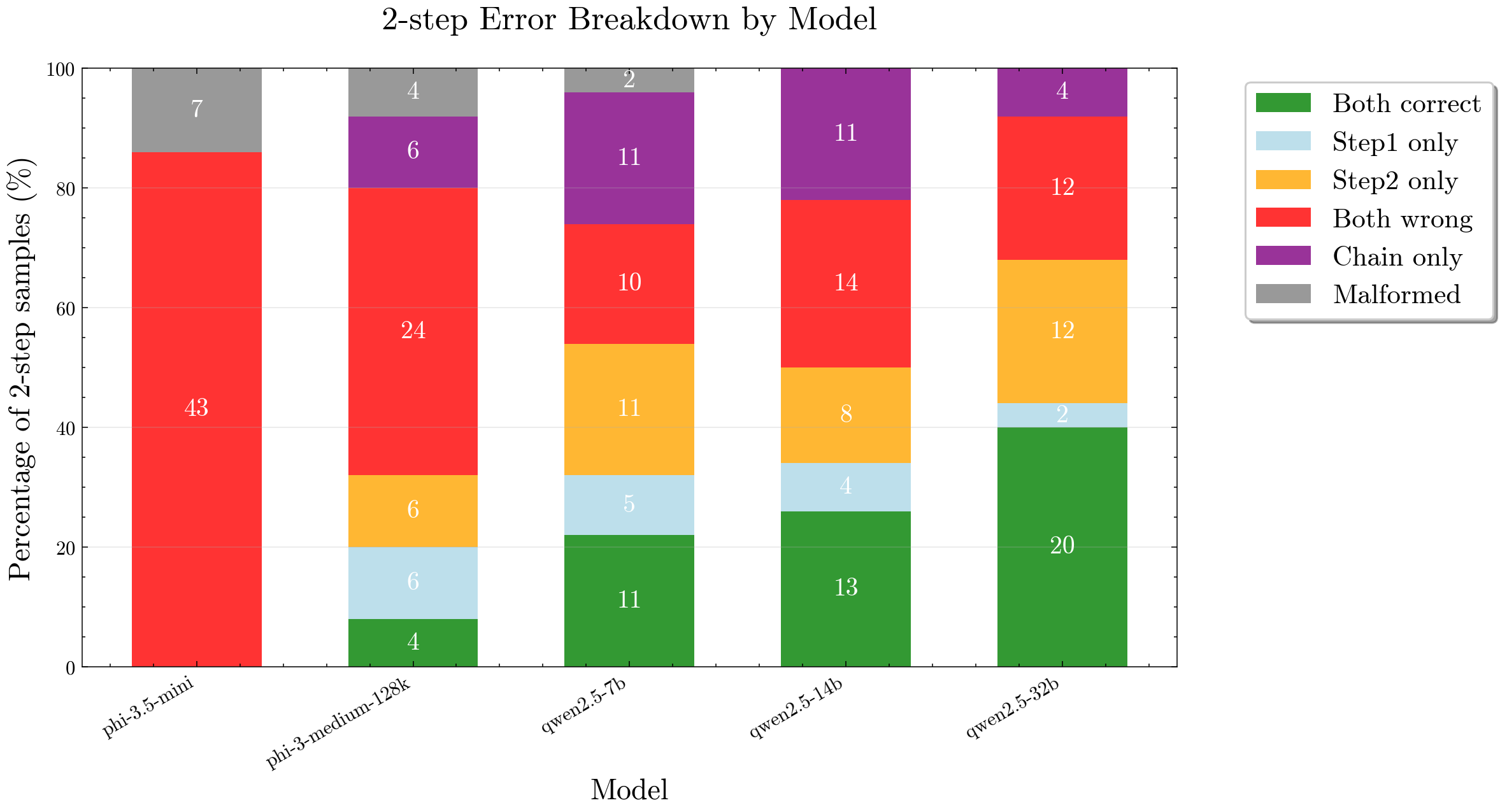}
  \caption{Error-type breakdown for the 2-step completion task by model. 
  Bars show the percentage of 2-step samples falling into each error category: \textit{Both correct}, \textit{Step1 only}, \textit{Step2 only}, \textit{Both wrong}, \textit{Chain only}, and \textit{Malformed}.}
  \label{fig:error-breakdown}
\end{figure}

Figure~\ref{fig:error-breakdown} shows the error breakdown by model. Most errors stem from step 1 (blue), step 2 (yellow), or both (red). Additional failures include formatting issues (grey, where the generated FOL expression was malformed and unparseable by SymPy) and chain-only errors, where the model captures general, overarching logical relations of the chain but introduces structural mistakes such as subexpression swaps or misplaced predicates. An example of chain-only errors are shown in Table~\ref{tab:step_completion_examples}. 

Our analyses indicate that reasoning LLMs may not be reasoning as rigorously as expected. They still perform poorly on both tasks 1 and 2, with error patterns revealing fundamental misunderstandings of basic logical syntax.

\section{Conclusion}

We introduce \textsc{FOL-Traces}, the first large-scale dataset of programmatically verified reasoning traces. In contrast to unverifiable natural language CoT corpora and limited theorem-proving datasets, \textsc{FOL-Traces} provides 7.4M verified examples with fine-grained complexity annotations, addressing a critical gap in reasoning evaluation and training. To assess its utility, we conduct pretraining sanity checks confirming stable optimization and coherent logical representations, followed by two structured reasoning diagnostics: \textit{masked operation prediction} and \textit{step completion}. Qwen2.5-32B attains 75\% accuracy on component prediction, while reasoning-tuned models such as Phi-3.5-mini perform markedly worse; on two-step completions, all models remain below 50\%. These results expose persistent limitations in compositional and process-level reasoning, establishing \textsc{FOL-Traces} as a scalable and verifiable benchmark for advancing genuine reasoning competence beyond surface-level pattern matching.

\section{Limitations and Future Work}

While \textsc{FOL-Traces} offers a transparent and verifiable foundation for studying structured reasoning in language models, several limitations remain. The dataset focuses on first-order logic, which, though expressive and formally grounded, captures only part of the reasoning diversity found in naturalistic contexts. Its symbolic generation process, while rigorously validated, diverges from the distributional properties of typical pretraining corpora, potentially limiting transferability to open-domain reasoning. Moreover, our evaluations target diagnostic reasoning tasks rather than full downstream applications, leaving generalization and scaling behavior for future study. Despite these constraints, \textsc{FOL-Traces} provides a controlled platform for probing symbolic competence and advancing more interpretable and transparent reasoning systems.

\section{Risks}

From a risk perspective, \textsc{FOL-Traces} would pose limited direct societal or ethical concerns, as it consists entirely of synthetic, abstract logical inferencing data and contains no personal or sensitive content. However, the use of LLMs for instantiation process may introduce subtle biases from the underlying models and their internet-scale data. Additionally, we contend that if our symbolic reasoning scope is not properly acknowledged in future work, strong performance on our dataset could be misinterpreted as evidence of broadly reliable or human-like reasoning if results are extrapolated beyond.

\section{Use of Generative AI}

We disclose generative AI usage in the preparation of this work. We used LLMs primarily as a writing assistant for correcting grammar and phrasing, as well as some related work suggestions. Additionally, LLMs were used for aiding in the dataset generation process as described in Section 2.2. We also evaluated current reasoning LLMs as described in Section 4. 

\bibliography{references}

@misc{Karpathy2022,
  author = {Andrej Karpathy},
  title = {\text{NanoGPT}},
  year = {2022},
  publisher = {GitHub},
  journal = {GitHub repository},
  howpublished = {\url{https://github.com/karpathy/nanoGPT}},
  commit = {325be85d9be8c81b436728a420e85796c57dba7e}
}

@misc{cobbe2021gsm8k,
      title={Training Verifiers to Solve Math Word Problems}, 
      author={Karl Cobbe and Vineet Kosaraju and Mohammad Bavarian and Mark Chen and Heewoo Jun and Lukasz Kaiser and Matthias Plappert and Jerry Tworek and Jacob Hilton and Reiichiro Nakano and Christopher Hesse and John Schulman},
      year={2021},
      eprint={2110.14168},
      archivePrefix={arXiv},
      primaryClass={cs.LG},
      url={https://arxiv.org/abs/2110.14168}, 
}

@article{hendrycksmath2021,
  title={Measuring Mathematical Problem Solving With the MATH Dataset},
  author={Dan Hendrycks and Collin Burns and Saurav Kadavath and Akul Arora and Steven Basart and Eric Tang and Dawn Song and Jacob Steinhardt},
  journal={NeurIPS},
  year={2021}
}

@misc{yu2024metamathbootstrapmathematicalquestions,
      title={MetaMath: Bootstrap Your Own Mathematical Questions for Large Language Models}, 
      author={Longhui Yu and Weisen Jiang and Han Shi and Jincheng Yu and Zhengying Liu and Yu Zhang and James T. Kwok and Zhenguo Li and Adrian Weller and Weiyang Liu},
      year={2024},
      eprint={2309.12284},
      archivePrefix={arXiv},
      primaryClass={cs.CL},
      url={https://arxiv.org/abs/2309.12284}, 
}

@misc{abdin2024phi3technicalreporthighly,
      title={Phi-3 Technical Report: A Highly Capable Language Model Locally on Your Phone}, 
      author={Marah Abdin and Jyoti Aneja and Hany Awadalla and Ahmed Awadallah and Ammar Ahmad Awan and Nguyen Bach and Amit Bahree and Arash Bakhtiari and Jianmin Bao and Harkirat Behl and Alon Benhaim and Misha Bilenko and Johan Bjorck and Sébastien Bubeck and Martin Cai and Qin Cai and Vishrav Chaudhary and Dong Chen and Dongdong Chen and Weizhu Chen and Yen-Chun Chen and Yi-Ling Chen and Hao Cheng and Parul Chopra and Xiyang Dai and Matthew Dixon and Ronen Eldan and Victor Fragoso and Jianfeng Gao and Mei Gao and Min Gao and Amit Garg and Allie Del Giorno and Abhishek Goswami and Suriya Gunasekar and Emman Haider and Junheng Hao and Russell J. Hewett and Wenxiang Hu and Jamie Huynh and Dan Iter and Sam Ade Jacobs and Mojan Javaheripi and Xin Jin and Nikos Karampatziakis and Piero Kauffmann and Mahoud Khademi and Dongwoo Kim and Young Jin Kim and Lev Kurilenko and James R. Lee and Yin Tat Lee and Yuanzhi Li and Yunsheng Li and Chen Liang and Lars Liden and Xihui Lin and Zeqi Lin and Ce Liu and Liyuan Liu and Mengchen Liu and Weishung Liu and Xiaodong Liu and Chong Luo and Piyush Madan and Ali Mahmoudzadeh and David Majercak and Matt Mazzola and Caio César Teodoro Mendes and Arindam Mitra and Hardik Modi and Anh Nguyen and Brandon Norick and Barun Patra and Daniel Perez-Becker and Thomas Portet and Reid Pryzant and Heyang Qin and Marko Radmilac and Liliang Ren and Gustavo de Rosa and Corby Rosset and Sambudha Roy and Olatunji Ruwase and Olli Saarikivi and Amin Saied and Adil Salim and Michael Santacroce and Shital Shah and Ning Shang and Hiteshi Sharma and Yelong Shen and Swadheen Shukla and Xia Song and Masahiro Tanaka and Andrea Tupini and Praneetha Vaddamanu and Chunyu Wang and Guanhua Wang and Lijuan Wang and Shuohang Wang and Xin Wang and Yu Wang and Rachel Ward and Wen Wen and Philipp Witte and Haiping Wu and Xiaoxia Wu and Michael Wyatt and Bin Xiao and Can Xu and Jiahang Xu and Weijian Xu and Jilong Xue and Sonali Yadav and Fan Yang and Jianwei Yang and Yifan Yang and Ziyi Yang and Donghan Yu and Lu Yuan and Chenruidong Zhang and Cyril Zhang and Jianwen Zhang and Li Lyna Zhang and Yi Zhang and Yue Zhang and Yunan Zhang and Xiren Zhou},
      year={2024},
      eprint={2404.14219},
      archivePrefix={arXiv},
      primaryClass={cs.CL},
      url={https://arxiv.org/abs/2404.14219}, 
}

@misc{dong2025reinforcementpretraining,
      title={Reinforcement Pre-Training}, 
      author={Qingxiu Dong and Li Dong and Yao Tang and Tianzhu Ye and Yutao Sun and Zhifang Sui and Furu Wei},
      year={2025},
      eprint={2506.08007},
      archivePrefix={arXiv},
      primaryClass={cs.CL},
      url={https://arxiv.org/abs/2506.08007}, 
}

@misc{hatamizadeh2025rlpreinforcementpretrainingobjective,
      title={RLP: Reinforcement as a Pretraining Objective}, 
      author={Ali Hatamizadeh and Syeda Nahida Akter and Shrimai Prabhumoye and Jan Kautz and Mostofa Patwary and Mohammad Shoeybi and Bryan Catanzaro and Yejin Choi},
      year={2025},
      eprint={2510.01265},
      archivePrefix={arXiv},
      primaryClass={cs.LG},
      url={https://arxiv.org/abs/2510.01265}, 
}

@inproceedings{Yang2023LeanDojoTP,
  title={{LeanDojo}: Theorem Proving with Retrieval-Augmented Language Models},
  author={Yang, Kaiyu and Swope, Aidan and Gu, Alex and Chalamala, Rahul and Song, Peiyang and Yu, Shixing and Godil, Saad and Prenger, Ryan and Anandkumar, Anima},
  booktitle={Neural Information Processing Systems (NeurIPS)},
  year={2023}
}

@inproceedings{lyu2023faithfulchainofthoughtreasoning,
    title = "Faithful Chain-of-Thought Reasoning",
    author = "Lyu, Qing  and
      Havaldar, Shreya  and
      Stein, Adam  and
      Zhang, Li  and
      Rao, Delip  and
      Wong, Eric  and
      Apidianaki, Marianna  and
      Callison-Burch, Chris",
    editor = "Park, Jong C.  and
      Arase, Yuki  and
      Hu, Baotian  and
      Lu, Wei  and
      Wijaya, Derry  and
      Purwarianti, Ayu  and
      Krisnadhi, Adila Alfa",
    booktitle = "Proceedings of the 13th International Joint Conference on Natural Language Processing and the 3rd Conference of the Asia-Pacific Chapter of the Association for Computational Linguistics (Volume 1: Long Papers)",
    month = nov,
    year = "2023",
    address = "Nusa Dua, Bali",
    publisher = "Association for Computational Linguistics",
    url = "https://aclanthology.org/2023.ijcnlp-main.20/",
    doi = "10.18653/v1/2023.ijcnlp-main.20",
    pages = "305--329"
}

@misc{wang2023selfconsistencyimproveschainthought,
      title={Self-Consistency Improves Chain of Thought Reasoning in Language Models}, 
      author={Xuezhi Wang and Jason Wei and Dale Schuurmans and Quoc Le and Ed Chi and Sharan Narang and Aakanksha Chowdhery and Denny Zhou},
      year={2023},
      eprint={2203.11171},
      archivePrefix={arXiv},
      primaryClass={cs.CL},
      url={https://arxiv.org/abs/2203.11171}, 
}

@misc{chia2023contrastivechainofthoughtprompting,
      title={Contrastive Chain-of-Thought Prompting}, 
      author={Yew Ken Chia and Guizhen Chen and Luu Anh Tuan and Soujanya Poria and Lidong Bing},
      year={2023},
      eprint={2311.09277},
      archivePrefix={arXiv},
      primaryClass={cs.CL},
      url={https://arxiv.org/abs/2311.09277}, 
}

@inproceedings{Wei_Jie_2024,
    title = "How Interpretable are Reasoning Explanations from Prompting Large Language Models?",
    author = "Wei Jie, Yeo  and
      Satapathy, Ranjan  and
      Goh, Rick  and
      Cambria, Erik",
    editor = "Duh, Kevin  and
      Gomez, Helena  and
      Bethard, Steven",
    booktitle = "Findings of the Association for Computational Linguistics: NAACL 2024",
    month = jun,
    year = "2024",
    address = "Mexico City, Mexico",
    publisher = "Association for Computational Linguistics",
    url = "https://aclanthology.org/2024.findings-naacl.138/",
    doi = "10.18653/v1/2024.findings-naacl.138",
    pages = "2148--2164"
}

@misc{wang2022interpretabilitywildcircuitindirect,
      title={Interpretability in the Wild: a Circuit for Indirect Object Identification in GPT-2 small}, 
      author={Kevin Wang and Alexandre Variengien and Arthur Conmy and Buck Shlegeris and Jacob Steinhardt},
      year={2022},
      eprint={2211.00593},
      archivePrefix={arXiv},
      primaryClass={cs.LG},
      url={https://arxiv.org/abs/2211.00593}, 
}

@misc{kojima2023largelanguagemodelszeroshot,
      title={Large Language Models are Zero-Shot Reasoners}, 
      author={Takeshi Kojima and Shixiang Shane Gu and Machel Reid and Yutaka Matsuo and Yusuke Iwasawa},
      year={2023},
      eprint={2205.11916},
      archivePrefix={arXiv},
      primaryClass={cs.CL},
      url={https://arxiv.org/abs/2205.11916}, 
}

@misc{elhage2022toymodelssuperposition,
      title={Toy Models of Superposition}, 
      author={Nelson Elhage and Tristan Hume and Catherine Olsson and Nicholas Schiefer and Tom Henighan and Shauna Kravec and Zac Hatfield-Dodds and Robert Lasenby and Dawn Drain and Carol Chen and Roger Grosse and Sam McCandlish and Jared Kaplan and Dario Amodei and Martin Wattenberg and Christopher Olah},
      year={2022},
      eprint={2209.10652},
      archivePrefix={arXiv},
      primaryClass={cs.LG},
      url={https://arxiv.org/abs/2209.10652}, 
}

@inproceedings{parmar2024logicbenchsystematicevaluationlogical,
    title = "{L}ogic{B}ench: Towards Systematic Evaluation of Logical Reasoning Ability of Large Language Models",
    author = "Parmar, Mihir  and
      Patel, Nisarg  and
      Varshney, Neeraj  and
      Nakamura, Mutsumi  and
      Luo, Man  and
      Mashetty, Santosh  and
      Mitra, Arindam  and
      Baral, Chitta",
    editor = "Ku, Lun-Wei  and
      Martins, Andre  and
      Srikumar, Vivek",
    booktitle = "Proceedings of the 62nd Annual Meeting of the Association for Computational Linguistics (Volume 1: Long Papers)",
    month = aug,
    year = "2024",
    address = "Bangkok, Thailand",
    publisher = "Association for Computational Linguistics",
    url = "https://aclanthology.org/2024.acl-long.739/",
    doi = "10.18653/v1/2024.acl-long.739",
    pages = "13679--13707"
}

@inproceedings{han2024folionaturallanguagereasoning,
    title = "{FOLIO}: Natural Language Reasoning with First-Order Logic",
    author = "Han, Simeng  and
      Schoelkopf, Hailey  and
      Zhao, Yilun  and
      Qi, Zhenting  and
      Riddell, Martin  and
      Zhou, Wenfei  and
      Coady, James  and
      Peng, David  and
      Qiao, Yujie  and
      Benson, Luke  and
      Sun, Lucy  and
      Wardle-Solano, Alexander  and
      Szab{\'o}, Hannah  and
      Zubova, Ekaterina  and
      Burtell, Matthew  and
      Fan, Jonathan  and
      Liu, Yixin  and
      Wong, Brian  and
      Sailor, Malcolm  and
      Ni, Ansong  and
      Nan, Linyong  and
      Kasai, Jungo  and
      Yu, Tao  and
      Zhang, Rui  and
      Fabbri, Alexander  and
      Kryscinski, Wojciech Maciej  and
      Yavuz, Semih  and
      Liu, Ye  and
      Lin, Xi Victoria  and
      Joty, Shafiq  and
      Zhou, Yingbo  and
      Xiong, Caiming  and
      Ying, Rex  and
      Cohan, Arman  and
      Radev, Dragomir",
    editor = "Al-Onaizan, Yaser  and
      Bansal, Mohit  and
      Chen, Yun-Nung",
    booktitle = "Proceedings of the 2024 Conference on Empirical Methods in Natural Language Processing",
    month = nov,
    year = "2024",
    address = "Miami, Florida, USA",
    publisher = "Association for Computational Linguistics",
    url = "https://aclanthology.org/2024.emnlp-main.1229/",
    doi = "10.18653/v1/2024.emnlp-main.1229",
    pages = "22017--22031"
}

@inproceedings{
jimenez2024swebench,
title={{SWE}-bench: Can Language Models Resolve Real-world Github Issues?},
author={Carlos E Jimenez and John Yang and Alexander Wettig and Shunyu Yao and Kexin Pei and Ofir Press and Karthik R Narasimhan},
booktitle={The Twelfth International Conference on Learning Representations},
year={2024},
url={https://openreview.net/forum?id=VTF8yNQM66}
}

@misc{wei2023chainofthoughtpromptingelicitsreasoning,
      title={Chain-of-Thought Prompting Elicits Reasoning in Large Language Models}, 
      author={Jason Wei and Xuezhi Wang and Dale Schuurmans and Maarten Bosma and Brian Ichter and Fei Xia and Ed Chi and Quoc Le and Denny Zhou},
      year={2023},
      eprint={2201.11903},
      archivePrefix={arXiv},
      primaryClass={cs.CL},
      url={https://arxiv.org/abs/2201.11903}, 
}

@misc{chen2021codex,
      title={Evaluating Large Language Models Trained on Code}, 
      author={Mark Chen and Jerry Tworek and Heewoo Jun and Qiming Yuan and Henrique Ponde de Oliveira Pinto and Jared Kaplan and Harri Edwards and Yuri Burda and Nicholas Joseph and Greg Brockman and Alex Ray and Raul Puri and Gretchen Krueger and Michael Petrov and Heidy Khlaaf and Girish Sastry and Pamela Mishkin and Brooke Chan and Scott Gray and Nick Ryder and Mikhail Pavlov and Alethea Power and Lukasz Kaiser and Mohammad Bavarian and Clemens Winter and Philippe Tillet and Felipe Petroski Such and Dave Cummings and Matthias Plappert and Fotios Chantzis and Elizabeth Barnes and Ariel Herbert-Voss and William Hebgen Guss and Alex Nichol and Alex Paino and Nikolas Tezak and Jie Tang and Igor Babuschkin and Suchir Balaji and Shantanu Jain and William Saunders and Christopher Hesse and Andrew N. Carr and Jan Leike and Josh Achiam and Vedant Misra and Evan Morikawa and Alec Radford and Matthew Knight and Miles Brundage and Mira Murati and Katie Mayer and Peter Welinder and Bob McGrew and Dario Amodei and Sam McCandlish and Ilya Sutskever and Wojciech Zaremba},
      year={2021},
      eprint={2107.03374},
      archivePrefix={arXiv},
      primaryClass={cs.LG},
      url={https://arxiv.org/abs/2107.03374}, 
}

@misc{geiping2025scalingtesttimecomputelatent,
      title={Scaling up Test-Time Compute with Latent Reasoning: A Recurrent Depth Approach}, 
      author={Jonas Geiping and Sean McLeish and Neel Jain and John Kirchenbauer and Siddharth Singh and Brian R. Bartoldson and Bhavya Kailkhura and Abhinav Bhatele and Tom Goldstein},
      year={2025},
      eprint={2502.05171},
      archivePrefix={arXiv},
      primaryClass={cs.LG},
      url={https://arxiv.org/abs/2502.05171}, 
}

@misc{vals2025aime,
  author       = {ValsAI},
  title        = {AIME Benchmark},
  year         = {2025},
  month        = mar,
  url          = {https://www.vals.ai/benchmarks/aime},
  note         = {Accessed: 2025-05-03}
}

@book{boolos2007computability, place={Cambridge}, edition={5}, title={Computability and Logic}, publisher={Cambridge University Press}, author={Boolos, George S. and Burgess, John P. and Jeffrey, Richard C.}, year={2007}}

@inproceedings{turpin2023languagemodelsdontsay,
author = {Turpin, Miles and Michael, Julian and Perez, Ethan and Bowman, Samuel R.},
title = {Language models don't always say what they think: unfaithful explanations in chain-of-thought prompting},
year = {2023},
publisher = {Curran Associates Inc.},
address = {Red Hook, NY, USA},
booktitle = {Proceedings of the 37th International Conference on Neural Information Processing Systems},
articleno = {3275},
numpages = {14},
location = {New Orleans, LA, USA},
series = {NIPS '23}
}

@misc{ruan2025reasoninglearnlatentthoughts,
      title={Reasoning to Learn from Latent Thoughts}, 
      author={Yangjun Ruan and Neil Band and Chris J. Maddison and Tatsunori Hashimoto},
      year={2025},
      eprint={2503.18866},
      archivePrefix={arXiv},
      primaryClass={cs.LG},
      url={https://arxiv.org/abs/2503.18866}, 
}

@misc{tack2025llmpretrainingcontinuousconcepts,
      title={LLM Pretraining with Continuous Concepts}, 
      author={Jihoon Tack and Jack Lanchantin and Jane Yu and Andrew Cohen and Ilia Kulikov and Janice Lan and Shibo Hao and Yuandong Tian and Jason Weston and Xian Li},
      year={2025},
      eprint={2502.08524},
      archivePrefix={arXiv},
      primaryClass={cs.LG},
      url={https://arxiv.org/abs/2502.08524}, 
}

@inproceedings{bender-koller-2020-climbing,
    title = "Climbing towards {NLU}: {On} Meaning, Form, and Understanding in the Age of Data",
    author = "Bender, Emily M.  and
      Koller, Alexander",
    editor = "Jurafsky, Dan  and
      Chai, Joyce  and
      Schluter, Natalie  and
      Tetreault, Joel",
    booktitle = "Proceedings of the 58th Annual Meeting of the Association for Computational Linguistics",
    month = jul,
    year = "2020",
    address = "Online",
    publisher = "Association for Computational Linguistics",
    url = "https://aclanthology.org/2020.acl-main.463/",
    doi = "10.18653/v1/2020.acl-main.463",
    pages = "5185--5198"
}

@inproceedings{min2022rethinkingroledemonstrationsmakes,
    title = "Rethinking the Role of Demonstrations: What Makes In-Context Learning Work?",
    author = "Min, Sewon  and
      Lyu, Xinxi  and
      Holtzman, Ari  and
      Artetxe, Mikel  and
      Lewis, Mike  and
      Hajishirzi, Hannaneh  and
      Zettlemoyer, Luke",
    editor = "Goldberg, Yoav  and
      Kozareva, Zornitsa  and
      Zhang, Yue",
    booktitle = "Proceedings of the 2022 Conference on Empirical Methods in Natural Language Processing",
    month = dec,
    year = "2022",
    address = "Abu Dhabi, United Arab Emirates",
    publisher = "Association for Computational Linguistics",
    url = "https://aclanthology.org/2022.emnlp-main.759/",
    doi = "10.18653/v1/2022.emnlp-main.759",
    pages = "11048--11064"
}

@inproceedings{tyen2024llmsreasoningerrorscorrect,
    title = "{LLM}s cannot find reasoning errors, but can correct them given the error location",
    author = "Tyen, Gladys  and
      Mansoor, Hassan  and
      Carbune, Victor  and
      Chen, Peter  and
      Mak, Tony",
    editor = "Ku, Lun-Wei  and
      Martins, Andre  and
      Srikumar, Vivek",
    booktitle = "Findings of the Association for Computational Linguistics: ACL 2024",
    month = aug,
    year = "2024",
    address = "Bangkok, Thailand",
    publisher = "Association for Computational Linguistics",
    url = "https://aclanthology.org/2024.findings-acl.826/",
    doi = "10.18653/v1/2024.findings-acl.826",
    pages = "13894--13908"
}

@article{lindsey2025biology,
  author={Lindsey, Jack and Gurnee, Wes and Ameisen, Emmanuel and Chen, Brian and Pearce, Adam and Turner, Nicholas L. and Citro, Craig and Abrahams, David and Carter, Shan and Hosmer, Basil and Marcus, Jonathan and Sklar, Michael and Templeton, Adly and Bricken, Trenton and McDougall, Callum and Cunningham, Hoagy and Henighan, Thomas and Jermyn, Adam and Jones, Andy and Persic, Andrew and Qi, Zhenyi and Thompson, T. Ben and Zimmerman, Sam and Rivoire, Kelley and Conerly, Thomas and Olah, Chris and Batson, Joshua},
  title={On the Biology of a Large Language Model},
  journal={Transformer Circuits Thread},
  year={2025},
  url={https://transformer-circuits.pub/2025/attribution-graphs/biology.html}
}

@misc{openai2024o1,
      title={OpenAI o1 System Card}, 
      author={OpenAI and : and Aaron Jaech and Adam Kalai and Adam Lerer and Adam Richardson and Ahmed El-Kishky and Aiden Low and Alec Helyar and Aleksander Madry and Alex Beutel and Alex Carney and Alex Iftimie and Alex Karpenko and Alex Tachard Passos and Alexander Neitz and Alexander Prokofiev and Alexander Wei and Allison Tam and Ally Bennett and Ananya Kumar and Andre Saraiva and Andrea Vallone and Andrew Duberstein and Andrew Kondrich and Andrey Mishchenko and Andy Applebaum and Angela Jiang and Ashvin Nair and Barret Zoph and Behrooz Ghorbani and Ben Rossen and Benjamin Sokolowsky and Boaz Barak and Bob McGrew and Borys Minaiev and Botao Hao and Bowen Baker and Brandon Houghton and Brandon McKinzie and Brydon Eastman and Camillo Lugaresi and Cary Bassin and Cary Hudson and Chak Ming Li and Charles de Bourcy and Chelsea Voss and Chen Shen and Chong Zhang and Chris Koch and Chris Orsinger and Christopher Hesse and Claudia Fischer and Clive Chan and Dan Roberts and Daniel Kappler and Daniel Levy and Daniel Selsam and David Dohan and David Farhi and David Mely and David Robinson and Dimitris Tsipras and Doug Li and Dragos Oprica and Eben Freeman and Eddie Zhang and Edmund Wong and Elizabeth Proehl and Enoch Cheung and Eric Mitchell and Eric Wallace and Erik Ritter and Evan Mays and Fan Wang and Felipe Petroski Such and Filippo Raso and Florencia Leoni and Foivos Tsimpourlas and Francis Song and Fred von Lohmann and Freddie Sulit and Geoff Salmon and Giambattista Parascandolo and Gildas Chabot and Grace Zhao and Greg Brockman and Guillaume Leclerc and Hadi Salman and Haiming Bao and Hao Sheng and Hart Andrin and Hessam Bagherinezhad and Hongyu Ren and Hunter Lightman and Hyung Won Chung and Ian Kivlichan and Ian O'Connell and Ian Osband and Ignasi Clavera Gilaberte and Ilge Akkaya and Ilya Kostrikov and Ilya Sutskever and Irina Kofman and Jakub Pachocki and James Lennon and Jason Wei and Jean Harb and Jerry Twore and Jiacheng Feng and Jiahui Yu and Jiayi Weng and Jie Tang and Jieqi Yu and Joaquin Quiñonero Candela and Joe Palermo and Joel Parish and Johannes Heidecke and John Hallman and John Rizzo and Jonathan Gordon and Jonathan Uesato and Jonathan Ward and Joost Huizinga and Julie Wang and Kai Chen and Kai Xiao and Karan Singhal and Karina Nguyen and Karl Cobbe and Katy Shi and Kayla Wood and Kendra Rimbach and Keren Gu-Lemberg and Kevin Liu and Kevin Lu and Kevin Stone and Kevin Yu and Lama Ahmad and Lauren Yang and Leo Liu and Leon Maksin and Leyton Ho and Liam Fedus and Lilian Weng and Linden Li and Lindsay McCallum and Lindsey Held and Lorenz Kuhn and Lukas Kondraciuk and Lukasz Kaiser and Luke Metz and Madelaine Boyd and Maja Trebacz and Manas Joglekar and Mark Chen and Marko Tintor and Mason Meyer and Matt Jones and Matt Kaufer and Max Schwarzer and Meghan Shah and Mehmet Yatbaz and Melody Y. Guan and Mengyuan Xu and Mengyuan Yan and Mia Glaese and Mianna Chen and Michael Lampe and Michael Malek and Michele Wang and Michelle Fradin and Mike McClay and Mikhail Pavlov and Miles Wang and Mingxuan Wang and Mira Murati and Mo Bavarian and Mostafa Rohaninejad and Nat McAleese and Neil Chowdhury and Neil Chowdhury and Nick Ryder and Nikolas Tezak and Noam Brown and Ofir Nachum and Oleg Boiko and Oleg Murk and Olivia Watkins and Patrick Chao and Paul Ashbourne and Pavel Izmailov and Peter Zhokhov and Rachel Dias and Rahul Arora and Randall Lin and Rapha Gontijo Lopes and Raz Gaon and Reah Miyara and Reimar Leike and Renny Hwang and Rhythm Garg and Robin Brown and Roshan James and Rui Shu and Ryan Cheu and Ryan Greene and Saachi Jain and Sam Altman and Sam Toizer and Sam Toyer and Samuel Miserendino and Sandhini Agarwal and Santiago Hernandez and Sasha Baker and Scott McKinney and Scottie Yan and Shengjia Zhao and Shengli Hu and Shibani Santurkar and Shraman Ray Chaudhuri and Shuyuan Zhang and Siyuan Fu and Spencer Papay and Steph Lin and Suchir Balaji and Suvansh Sanjeev and Szymon Sidor and Tal Broda and Aidan Clark and Tao Wang and Taylor Gordon and Ted Sanders and Tejal Patwardhan and Thibault Sottiaux and Thomas Degry and Thomas Dimson and Tianhao Zheng and Timur Garipov and Tom Stasi and Trapit Bansal and Trevor Creech and Troy Peterson and Tyna Eloundou and Valerie Qi and Vineet Kosaraju and Vinnie Monaco and Vitchyr Pong and Vlad Fomenko and Weiyi Zheng and Wenda Zhou and Wes McCabe and Wojciech Zaremba and Yann Dubois and Yinghai Lu and Yining Chen and Young Cha and Yu Bai and Yuchen He and Yuchen Zhang and Yunyun Wang and Zheng Shao and Zhuohan Li},
      year={2024},
      eprint={2412.16720},
      archivePrefix={arXiv},
      primaryClass={cs.AI},
      url={https://arxiv.org/abs/2412.16720}, 
}

@misc{bai2023qwentechnicalreport,
      title={Qwen Technical Report}, 
      author={Jinze Bai and Shuai Bai and Yunfei Chu and Zeyu Cui and Kai Dang and Xiaodong Deng and Yang Fan and Wenbin Ge and Yu Han and Fei Huang and Binyuan Hui and Luo Ji and Mei Li and Junyang Lin and Runji Lin and Dayiheng Liu and Gao Liu and Chengqiang Lu and Keming Lu and Jianxin Ma and Rui Men and Xingzhang Ren and Xuancheng Ren and Chuanqi Tan and Sinan Tan and Jianhong Tu and Peng Wang and Shijie Wang and Wei Wang and Shengguang Wu and Benfeng Xu and Jin Xu and An Yang and Hao Yang and Jian Yang and Shusheng Yang and Yang Yao and Bowen Yu and Hongyi Yuan and Zheng Yuan and Jianwei Zhang and Xingxuan Zhang and Yichang Zhang and Zhenru Zhang and Chang Zhou and Jingren Zhou and Xiaohuan Zhou and Tianhang Zhu},
      year={2023},
      eprint={2309.16609},
      archivePrefix={arXiv},
      primaryClass={cs.CL},
      url={https://arxiv.org/abs/2309.16609}, 
}

@article{deepseekai2025deepseekr1incentivizingreasoningcapability,
   title={DeepSeek-R1 incentivizes reasoning in LLMs through reinforcement learning},
   volume={645},
   ISSN={1476-4687},
   url={http://dx.doi.org/10.1038/s41586-025-09422-z},
   DOI={10.1038/s41586-025-09422-z},
   number={8081},
   journal={Nature},
   publisher={Springer Science and Business Media LLC},
   author={Guo, Daya and Yang, Dejian and Zhang, Haowei and Song, Junxiao and Wang, Peiyi and Zhu, Qihao and Xu, Runxin and Zhang, Ruoyu and Ma, Shirong and Bi, Xiao and Zhang, Xiaokang and Yu, Xingkai and Wu, Yu and Wu, Z. F. and Gou, Zhibin and Shao, Zhihong and Li, Zhuoshu and Gao, Ziyi and Liu, Aixin and Xue, Bing and Wang, Bingxuan and Wu, Bochao and Feng, Bei and Lu, Chengda and Zhao, Chenggang and Deng, Chengqi and Ruan, Chong and Dai, Damai and Chen, Deli and Ji, Dongjie and Li, Erhang and Lin, Fangyun and Dai, Fucong and Luo, Fuli and Hao, Guangbo and Chen, Guanting and Li, Guowei and Zhang, H. and Xu, Hanwei and Ding, Honghui and Gao, Huazuo and Qu, Hui and Li, Hui and Guo, Jianzhong and Li, Jiashi and Chen, Jingchang and Yuan, Jingyang and Tu, Jinhao and Qiu, Junjie and Li, Junlong and Cai, J. L. and Ni, Jiaqi and Liang, Jian and Chen, Jin and Dong, Kai and Hu, Kai and You, Kaichao and Gao, Kaige and Guan, Kang and Huang, Kexin and Yu, Kuai and Wang, Lean and Zhang, Lecong and Zhao, Liang and Wang, Litong and Zhang, Liyue and Xu, Lei and Xia, Leyi and Zhang, Mingchuan and Zhang, Minghua and Tang, Minghui and Zhou, Mingxu and Li, Meng and Wang, Miaojun and Li, Mingming and Tian, Ning and Huang, Panpan and Zhang, Peng and Wang, Qiancheng and Chen, Qinyu and Du, Qiushi and Ge, Ruiqi and Zhang, Ruisong and Pan, Ruizhe and Wang, Runji and Chen, R. J. and Jin, R. L. and Chen, Ruyi and Lu, Shanghao and Zhou, Shangyan and Chen, Shanhuang and Ye, Shengfeng and Wang, Shiyu and Yu, Shuiping and Zhou, Shunfeng and Pan, Shuting and Li, S. S. and Zhou, Shuang and Wu, Shaoqing and Yun, Tao and Pei, Tian and Sun, Tianyu and Wang, T. and Zeng, Wangding and Liu, Wen and Liang, Wenfeng and Gao, Wenjun and Yu, Wenqin and Zhang, Wentao and Xiao, W. L. and An, Wei and Liu, Xiaodong and Wang, Xiaohan and Chen, Xiaokang and Nie, Xiaotao and Cheng, Xin and Liu, Xin and Xie, Xin and Liu, Xingchao and Yang, Xinyu and Li, Xinyuan and Su, Xuecheng and Lin, Xuheng and Li, X. Q. and Jin, Xiangyue and Shen, Xiaojin and Chen, Xiaosha and Sun, Xiaowen and Wang, Xiaoxiang and Song, Xinnan and Zhou, Xinyi and Wang, Xianzu and Shan, Xinxia and Li, Y. K. and Wang, Y. Q. and Wei, Y. X. and Zhang, Yang and Xu, Yanhong and Li, Yao and Zhao, Yao and Sun, Yaofeng and Wang, Yaohui and Yu, Yi and Zhang, Yichao and Shi, Yifan and Xiong, Yiliang and He, Ying and Piao, Yishi and Wang, Yisong and Tan, Yixuan and Ma, Yiyang and Liu, Yiyuan and Guo, Yongqiang and Ou, Yuan and Wang, Yuduan and Gong, Yue and Zou, Yuheng and He, Yujia and Xiong, Yunfan and Luo, Yuxiang and You, Yuxiang and Liu, Yuxuan and Zhou, Yuyang and Zhu, Y. X. and Huang, Yanping and Li, Yaohui and Zheng, Yi and Zhu, Yuchen and Ma, Yunxian and Tang, Ying and Zha, Yukun and Yan, Yuting and Ren, Z. Z. and Ren, Zehui and Sha, Zhangli and Fu, Zhe and Xu, Zhean and Xie, Zhenda and Zhang, Zhengyan and Hao, Zhewen and Ma, Zhicheng and Yan, Zhigang and Wu, Zhiyu and Gu, Zihui and Zhu, Zijia and Liu, Zijun and Li, Zilin and Xie, Ziwei and Song, Ziyang and Pan, Zizheng and Huang, Zhen and Xu, Zhipeng and Zhang, Zhongyu and Zhang, Zhen},
   year={2025},
   month=sep, pages={633–638} }

\onecolumn
\appendix

\section{Appendix}
\subsection{Full Simplification Algorithm}
\label{sec:full-algo}

\begin{algorithm}
\caption{Get Depth of Logical Expression}
\begin{algorithmic}[1]
\Function{GetDepth}{expr}
    \If{expr is And, Or, or Implies}
        \State \Return $1 + \max($GetDepth of each subexpression$)$
    \ElsIf{expr is Not}
        \State \Return $1 +$ GetDepth of inner expression
    \Else
        \State \Return $0$
    \EndIf
\EndFunction
\end{algorithmic}
\end{algorithm}

\newpage
\begin{figure*}[!htbp]
  \centering
  \resizebox{\textwidth}{!}{%
    \parbox{\textwidth}{%
      \small
      \captionof{algorithm}{Simplify a Subexpression with Depth $<2$}
      \begin{algorithmic}[1]
        \Function{TraverseAndSimplify}{expr, d}
            \State complexity\_count $\gets$ complexity\_count + 1
            \If{simplified\_once} \Return expr \EndIf

            \If{GetDepth(expr) $<$ d}
                \For{$i \gets 0$ to $d - 1$}
                    \If{$i < 2$}
                        \State simplified\_expr $\gets$ SimplifyLogic(expr, deep=False)
                    \Else
                        \State simplified\_expr $\gets$ TraverseAndSimplify(expr, i)
                    \EndIf
                    \If{simplified\_expr $\neq$ expr}
                        \State simplified\_once $\gets$ True
                        \State \Return simplified\_expr
                    \EndIf
                \EndFor
            \EndIf

            \If{expr is Not}
                \If{expr.args[0] == True} \Return False \EndIf
                \If{expr.args[0] == False} \Return True \EndIf
                \If{expr.args[0] is Not} \Return expr.args[0].args[0] \EndIf
                \If{expr.args[0] is Or} \Return And(Not(x) for x in args) \EndIf
                \If{expr.args[0] is And} \Return Or(Not(x) for x in args) \EndIf
                \State \Return Not(TraverseAndSimplify(expr.args[0], d))
            \EndIf

            \If{expr is Implies}
                \If{expr.args[0] == expr.args[1]} \Return True \EndIf
                \If{both args are atoms and simplifiable} \Return simplified Implies \EndIf
                \If{GetDepth(expr) $<$ 5} \Return Or(Not(lhs), rhs) \EndIf
                \State \Return Implies(TraverseAndSimplify(lhs, d), TraverseAndSimplify(rhs, d))
            \EndIf

            \If{expr is And or Or}
                \For{arg in expr.args}
                    \If{GetDepth(arg) $<$ 2}
                        \State simplified $\gets$ SimplifyLogic(arg, deep=False)
                        \If{simplified $\neq$ arg}
                            \State \Return expr.func(...with simplified arg...)
                        \EndIf
                    \EndIf
                \EndFor
                \State \Return expr.func(TraverseAndSimplify(arg, d) for arg in expr.args)
            \EndIf

            \State \Return expr
        \EndFunction

        \Statex

        \Function{Main}{formula}
            \State simplified\_once $\gets$ False
            \State complexity\_count $\gets$ 0
            \State result $\gets$ TraverseAndSimplify(formula, d)
            \State \Return result, complexity\_count
        \EndFunction
      \end{algorithmic}
    }%
  }
\end{figure*}

\newpage
\section{Additional Dataset Distribution Information}
\begin{table}[!htbp]
\centering
\caption{Summary of bucket statistics.}
\label{tab:bucket_data}
\small
\begin{tabular}{c c c c c}
\toprule
Bucket & \#rules & Start & End & $\Delta$ \\
\midrule
2  & 82,490  & 15.59 & 5.76  & +9.83  \\
4  & 217,298 & 16.06 & 7.23  & +8.83  \\
8  & 225,177 & 17.42 & 9.61  & +7.82  \\
12 & 87,316  & 19.91 & 11.75 & +8.16  \\
15 & 25,925  & 22.44 & 13.65 & +8.79  \\
20 & 1,451   & 25.42 & 14.68 & +10.74 \\
25 & 15      & 28.00 & 13.80 & +14.20 \\
\bottomrule
\end{tabular}
\end{table}

\begin{figure}[!htbp]
    \centering
    \includegraphics[width=0.6\linewidth]{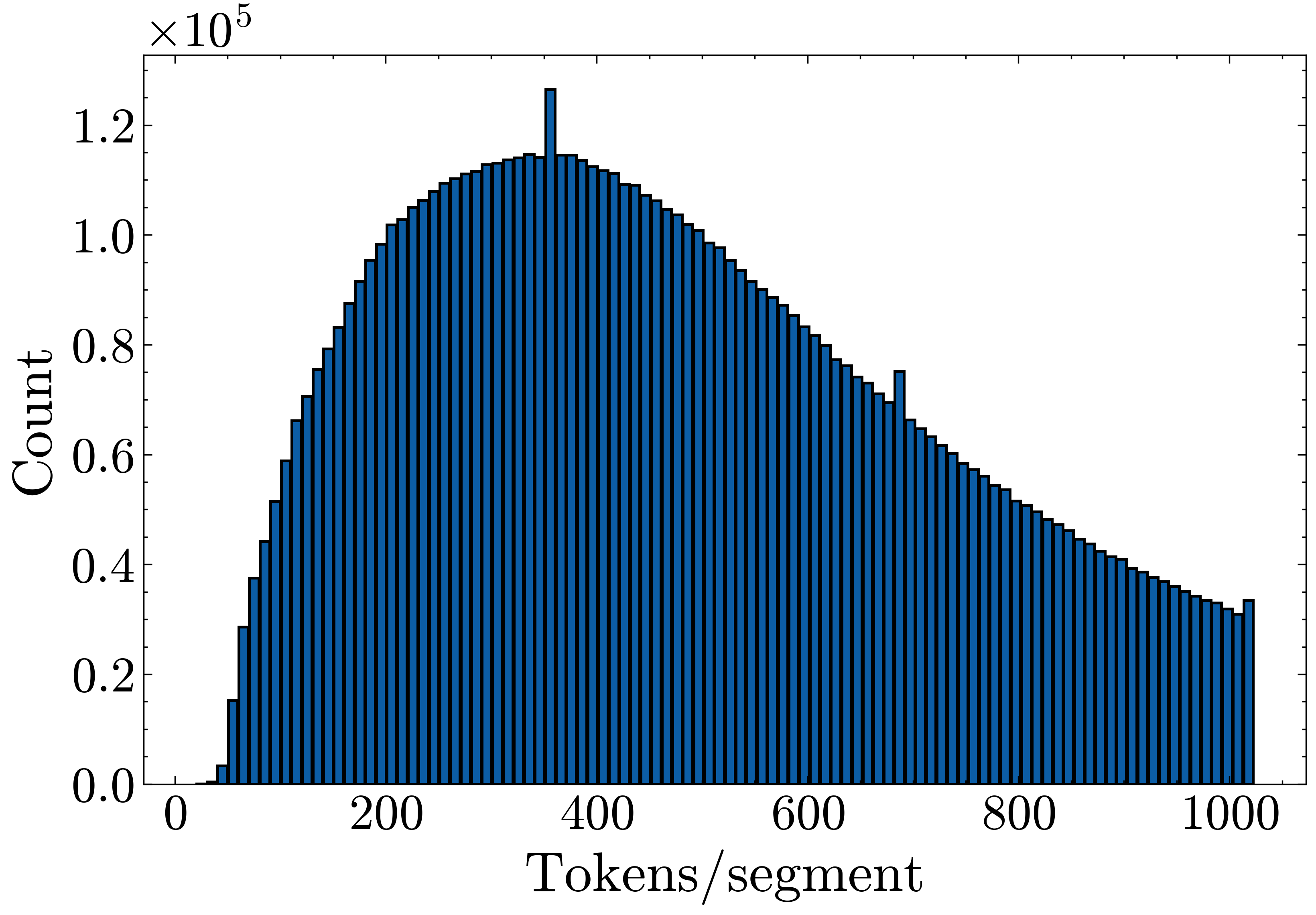}
    \caption{Training segment statistics. Distribution of token‐counts per segment.}
    \label{fig:train2_tokens}
\end{figure}




\begin{figure}[!htbp]
  \centering
  \begin{subfigure}[b]{0.32\textwidth}
    \includegraphics[width=\linewidth]{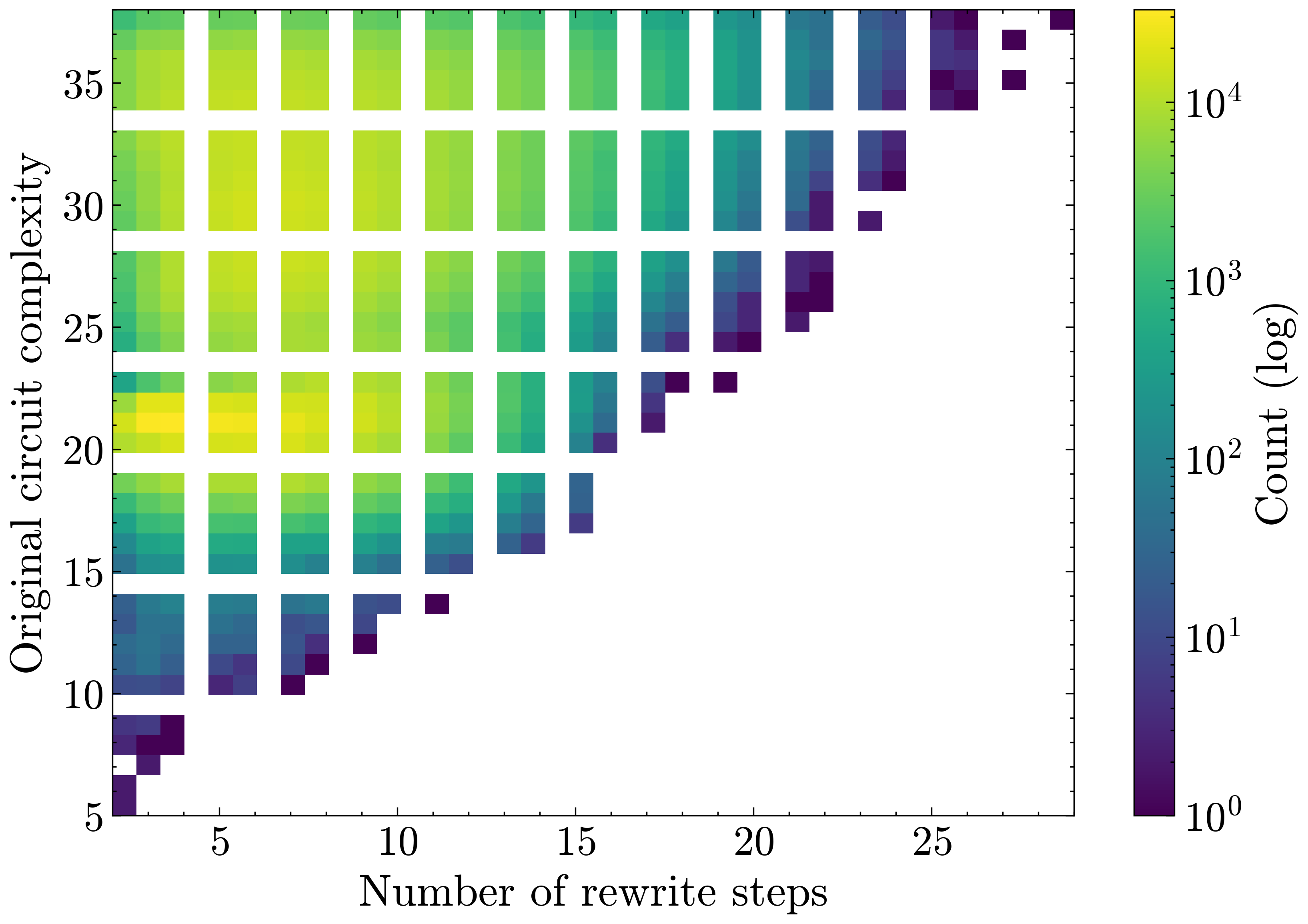}
    \caption{Log‐density heatmap of number of steps vs original complexity}
    \label{fig:steps_vs_orig_heatmap}
  \end{subfigure}\hfill
  \begin{subfigure}[b]{0.32\textwidth}
    \includegraphics[width=\linewidth]{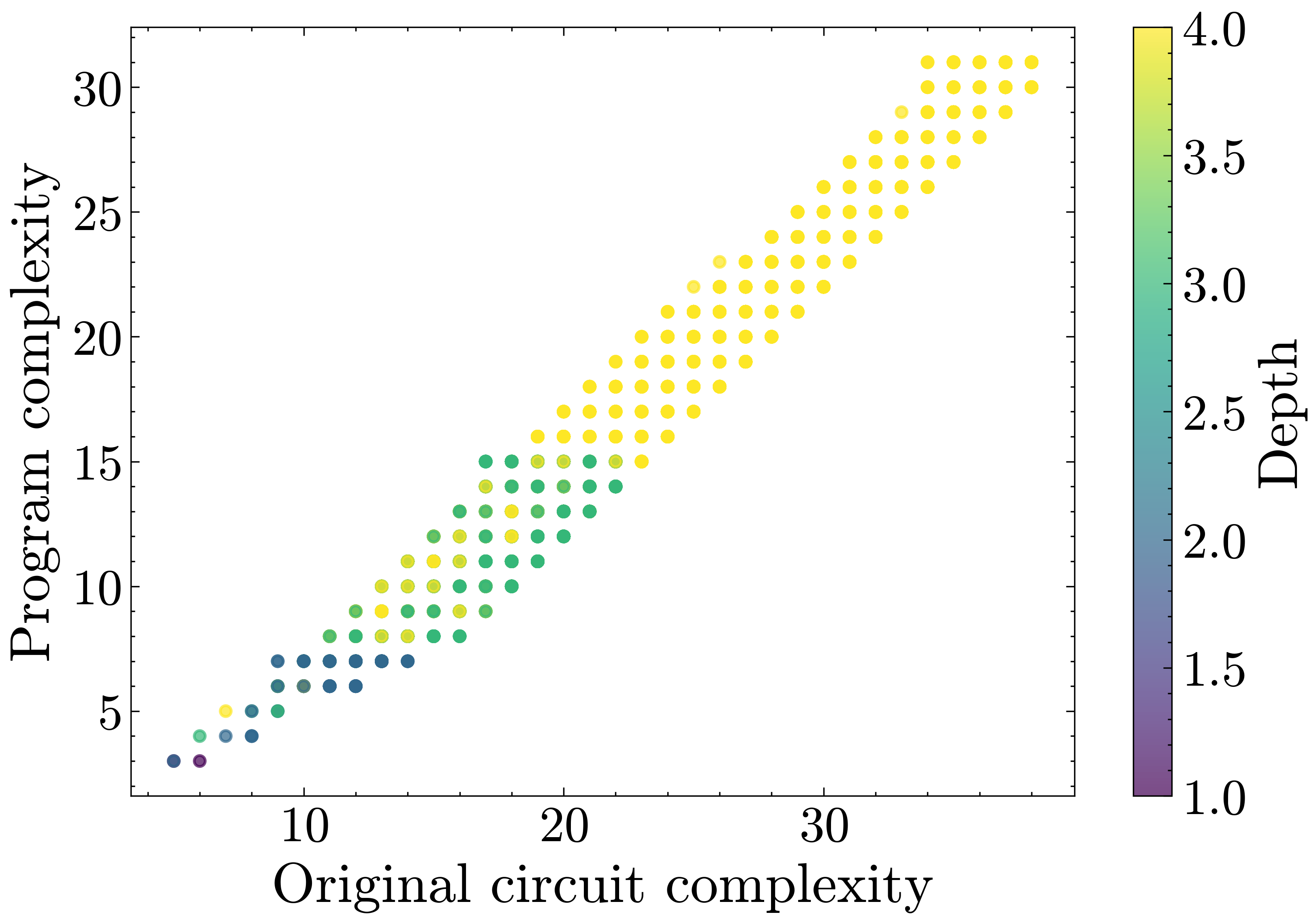}
    \caption{Scatter of original vs program complexity (colour=depth)}
    \label{fig:orig_vs_prog_scatter}
  \end{subfigure}\hfill
  \begin{subfigure}[b]{0.32\textwidth}
    \includegraphics[width=\linewidth, height=0.7\linewidth]{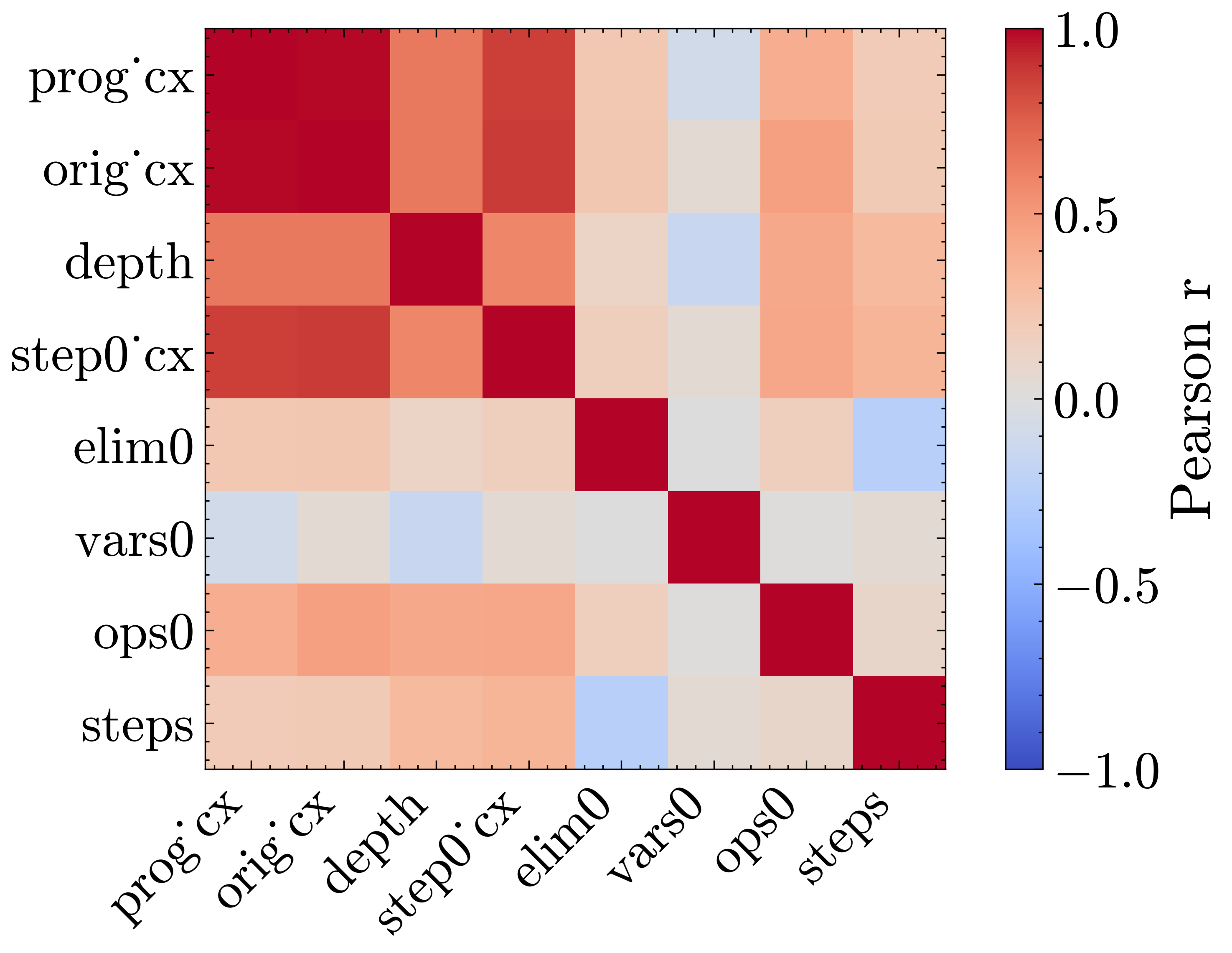}
    \caption{Pearson correlation matrix of rule‐level metrics}
    \label{fig:rule_corr_matrix}
  \end{subfigure}
  \caption{(a) Joint distribution of simplification steps and original circuit complexity;  
    (b) Relationship between original and program complexity, coloured by derivation depth;  
    (c) Correlations among key rule‐level features (circuit/program complexity, depth, vars, ops, etc.).}
  \label{fig:rule_diagnostics_combined}
\end{figure}

\newpage
\section{First Order Logic (FOL) Categories and Explanations}
\label{sec:fol-categories}
\begin{table*}[htbp]
\centering
\small
\scalebox{0.85}{%
\begin{tabular}{c c c}
\toprule
FOL Inference Rule & Symbolic Expression & Explanation \\
\midrule
Bidirectional Dilemma (BD) &
\(
\begin{array}{c}
((p \rightarrow q) \land (r \rightarrow s)), \\
(p \lor \neg s ) \models (q \lor \neg r)
\end{array}
\) &
\makecell[l]{If two conditional statements are true, given a true antecedent\\
or a false consequent, the respective consequent is true or\\
a respective antecedent is false.} \\
\midrule
Constructive Dilemma (CD) &
\(
\begin{array}{c}
((p \rightarrow q) \land (r \rightarrow s)), \\
(p \lor r ) \models (q \lor s)
\end{array}
\) &
\makecell[l]{If two conditional statements are true and at least one of\\
their antecedents is true, then at least one of their consequents is true.} \\
\midrule
Destructive Dilemma (DD) &
\(
\begin{array}{c}
((p \rightarrow q) \land (r \rightarrow s)), \\
(\neg q \lor \neg s ) \models (\neg p \lor \neg r)
\end{array}
\) &
\makecell[l]{If two conditional statements are true, and one of their consequents\\
must be false, then one of their antecedents must be false.} \\
\midrule
Disjunctive Syllogism (DS) &
\(
\begin{array}{c}
((p \lor q)\land \neg p)\models q
\end{array}
\) &
\makecell[l]{Disjunctive elimination. If we know one of two statements, \(p\) or \(q\),\\
to be true, and one of them is not true, the other must be true.} \\
\midrule
Hypothetical Syllogism (HS) &
\(
\begin{array}{c}
((p\rightarrow q)\land (q\rightarrow r)) \\
\models (p\rightarrow r)
\end{array}
\) &
\makecell[l]{Chain argument rule or transitivity of implication.} \\
\midrule
Modus Ponens (MP) &
\(
\begin{array}{c}
((p\rightarrow q)\land p)\models q
\end{array}
\) &
\makecell[l]{Implication elimination rule. If \(p\) implies \(q\) and \(p\) is true,\\
the statement can be replaced with \(q\).} \\
\midrule
Modus Tollens (MT) &
\(
\begin{array}{c}
((p\rightarrow q)\land \neg q)\models \neg p
\end{array}
\) &
\makecell[l]{Implication elimination rule. If \(p\) implies \(q\) and \(q\) is false,\\
the statement can be replaced with \(\neg p\).} \\
\midrule
Universal Instantiation (UI) &
\(
\begin{array}{c}
\forall x P(x)\implies \exists a P(a)
\end{array}
\) &
\makecell[l]{If a statement \(P\) holds for a variable \(x\), then there exists a\\
particular value \(a\) for the statement to be true.} \\
\midrule
Existential Generalization (EG) &
\(
\begin{array}{c}
\exists x P(x)\implies P(a)
\end{array}
\) &
\makecell[l]{If a statement \(P\) holds true for some subset of variables \(x\),\\
then there is a particular value \(x=a\) for which \(P\) holds true.}\\
\midrule
FOL proofs \& general statements & -- & -- \\
\bottomrule
\end{tabular}}
\captionsetup{skip=5pt}
\caption{First Order Logic (FOL) inference rule categories and explanations.\protect\footnotemark}
\label{tab:fol-inference-rules}
\end{table*}

\begin{table}[htpb]
\centering

\scalebox{0.80}{
\begin{tabular}{cc}
\toprule
 FOL Properties & Symbolic Expression  \\
 \midrule
 \midrule
 Distributive (Dist) & \makecell{
 \(
 \begin{aligned}
     &(p\lor (q\land r)) \leftrightarrow ((p\lor q)\land (p\lor r))\\
     &(p\land (q\lor r)) \leftrightarrow ((p\land q)\lor (p\land r))
 \end{aligned}
 \)
 } \\
 \midrule
 Association (AS) & \makecell{
 \(
 \begin{aligned}
     &(p\lor (q\lor r)) \leftrightarrow ((p\lor q) \lor r)\\
     &(p\land (q\land r)) \leftrightarrow ((p\land q) \land r)
 \end{aligned}
 \)
 } \\
 \midrule
 Tautology (TT) & \makecell{
 \(
 \begin{aligned}
     &p \leftrightarrow (p\lor p)\\
     &p \leftrightarrow (p\land p)
 \end{aligned}
 \)
 }\\
 \midrule
 Transposition (TS) & \makecell{
 \(
 \begin{aligned}
     (p\rightarrow q) \leftrightarrow (\neg q \rightarrow \neg p)
 \end{aligned}
 \)
 }\\
 \midrule
 Importation (IM) & \makecell{
 \(
 \begin{aligned}
     (p\rightarrow (q\rightarrow r)) \leftrightarrow ((p\land q)\rightarrow r)
 \end{aligned}
 \)
 }\\
 \midrule
 Exportation (EX) & \makecell{
 \(
 \begin{aligned}
     ((p\land q)\rightarrow r) \rightarrow (p\rightarrow (q\rightarrow r))
 \end{aligned}
 \)
 }\\
 \midrule
 Double Negation (DN) & \makecell{
 \(
 \begin{aligned}
     p\leftrightarrow \neg \neg p
 \end{aligned}
 \)
 }\\
 \midrule
 De Morgan's Law (DM) & \makecell{
 \(
 \begin{aligned}
     &\neg (p\land q) \leftrightarrow (\neg p\lor \neg q )\\
     &\neg (p\lor q) \leftrightarrow (\neg p \land \neg q)
 \end{aligned}
 \)
 }\\
 \midrule
 Negation of XOR (NX) & \makecell{
 \(
 \begin{aligned}
     &\neg (p\oplus q)\leftrightarrow (\neg p \oplus \neg q)\\
     &\neg (p\oplus q)\leftrightarrow (p\odot q)
 \end{aligned}
 \)
 }\\
 \midrule
 Negation of XNOR (NN) & \makecell{
 \(
 \begin{aligned}
     &\neg(p\odot q) \leftrightarrow (\neg p \odot \neg q)\\
     &\neg(p\odot q) \leftrightarrow (p \oplus q)\\
 \end{aligned}
 \)
 }\\
\bottomrule
\end{tabular}
}
\captionsetup{skip=5pt}
\caption{First Order Logic (FOL) Basic Properties\protect\footnotemark}
\label{tab:fol-basic}
\end{table}

\newpage
\section{Elimination Rules}
\label{sec:fol-eliminations-categories}
\begin{table}[ht]
    \centering
    \small
    \scalebox{0.80}{
    \begin{tabular}{cc}
        \toprule
         & Symbolic Expression \\ 
        \midrule
        \midrule
        E0  & \makecell[l]{
 $\begin{aligned}
     &p \lor \text{True} \leftrightarrow \text{True}
 \end{aligned}$} \\ \midrule
        E1  & 
        \makecell[l]{
 $\begin{aligned}
     &p \lor \text{False} \leftrightarrow p
 \end{aligned}$}\\ \midrule
        E2  & \makecell[l]{
 $\begin{aligned}
     &p \land \text{True} \leftrightarrow p
 \end{aligned}$} \\ \midrule
        E3  & \makecell[l]{
 $\begin{aligned}
     &p \land \text{False} \leftrightarrow \text{False}
 \end{aligned}$} \\ \midrule
        E4  & \makecell[l]{$\begin{aligned}
     &\text{True} \lor p \leftrightarrow \text{True}
 \end{aligned}$} \\ \midrule
        E5  & \makecell[l]{$\begin{aligned}
     &\text{False} \lor p \leftrightarrow p
 \end{aligned}$} \\ \midrule

        E6  & \makecell[l]{$\begin{aligned}      &\text{True} \land p \leftrightarrow p  \end{aligned}$} \\ \midrule
        E7  & \makecell[l]{$\begin{aligned}      &\text{False} \land p \leftrightarrow \text{False}  \end{aligned}$} \\ \midrule
        E8  & \makecell[l]{$\begin{aligned}      &p \lor p \leftrightarrow p  \end{aligned}$} \\ \midrule
        E9  & \makecell[l]{$\begin{aligned}      &p \land p \leftrightarrow p  \end{aligned}$} \\ \midrule
        E10 & \makecell[l]{$\begin{aligned}      &p \land \neg p \leftrightarrow \text{False}  \end{aligned}$} \\ \midrule
        E11 & \makecell[l]{$\begin{aligned}      &p \lor \neg p \leftrightarrow \text{True}  \end{aligned}$} \\ \midrule
        E12 & \makecell[l]{$\begin{aligned}      &\neg p \land p \leftrightarrow \text{False}  \end{aligned}$} \\ \midrule
        E13 & \makecell[l]{$\begin{aligned}      &\neg p \lor p \leftrightarrow \text{True}  \end{aligned}$} \\ \midrule
        E14 & \makecell[l]{$\begin{aligned}      &p \land (p \lor q) \leftrightarrow p  \end{aligned}$} \\ \midrule
        E15 & \makecell[l]{
 $\begin{aligned}
     &p \land (\neg p \lor q) \leftrightarrow (p \land \neg p) \lor (p \land q) \\
     & \leftrightarrow \text{False} \lor (p \land q) \leftrightarrow p \land q
 \end{aligned}$}\\ \midrule
        E16 &

        \makecell[l]{$\begin{aligned}      &p \land (\neg p \lor q) \leftrightarrow \text{False} \lor (p \land q) \leftrightarrow p \land q  \end{aligned}$} \\ \midrule
        E17 & \makecell[l]{$\begin{aligned}      &p \land (\neg p \lor q) \leftrightarrow (p \land \neg p) \lor (p \land q) \leftrightarrow p \land q  \end{aligned}$} \\ \midrule
        E18 & \makecell[l]{$\begin{aligned}      &p \lor (p \land q) \leftrightarrow p  \end{aligned}$} \\ \midrule
        E19 & \makecell[l]{$\begin{aligned}      &p \lor (p \land q \land r) \leftrightarrow p  \end{aligned}$} \\ \midrule
        E20 & \makecell[l]{$\begin{aligned}      &r \lor (p \land q \land r) \leftrightarrow r  \end{aligned}$} \\ \midrule
        E21 & \makecell[l]{$\begin{aligned}      &r \lor (p \land q \land r \land s) \leftrightarrow r  \end{aligned}$} \\ \midrule
        E22 & \makecell[l]{$\begin{aligned}      &p \lor (\neg p \land q) \leftrightarrow (p \lor \neg p) \land (p \lor q) \\
        &\leftrightarrow \text{True} \land (p \lor q) \leftrightarrow (p \lor q)  \end{aligned}$} \\ \midrule
        E23 & \makecell[l]{$\begin{aligned}      &p \lor (\neg p \land q) \leftrightarrow \text{True} \land (p \lor q) \leftrightarrow (p \lor q)  \end{aligned}$} \\ \midrule
        E24 & \makecell[l]{$\begin{aligned}      &p \lor (\neg p \land q) \leftrightarrow (p \lor \neg p) \land (p \lor q) \leftrightarrow (p \lor q)  \end{aligned}$} \\ \midrule
        E25 & \makecell[l]{
 $\begin{aligned}
     &p \lor \neg(p \land q) \leftrightarrow p \lor (\neg p \lor \neg q)\\
     &\leftrightarrow (p \lor \neg p) \lor \neg q \leftrightarrow \text{True} \lor \neg q \leftrightarrow \text{True}
 \end{aligned}$}\\ \midrule
        E26 & \makecell[l]{$\begin{aligned}      &p \lor \neg(p \land q) \leftrightarrow p \lor (\neg p \lor \neg q) \\
        &\leftrightarrow p \lor \neg p \lor \neg q \leftrightarrow \text{True} \lor \neg q \leftrightarrow \text{True}  \end{aligned}$} \\ \midrule
        E27 & \makecell[l]{$\begin{aligned}      &p \lor \neg(p \land q) \leftrightarrow (p \lor \neg p) \lor \neg q \leftrightarrow \text{True} \lor \neg q \leftrightarrow \text{True}  \end{aligned}$} \\ \midrule
        E28 & \makecell[l]{$\begin{aligned}      &p \lor \neg(p \land q) \leftrightarrow p \lor (\neg p \lor \neg q) \leftrightarrow \text{True} \lor \neg q \leftrightarrow \text{True}  \end{aligned}$} \\ \midrule
        E29 & \makecell[l]{$\begin{aligned}      &p \lor \neg(p \land q) \leftrightarrow p \lor (\neg p \lor \neg q) \leftrightarrow (p \lor \neg p) \lor \neg q \leftrightarrow \text{True}  \end{aligned}$} \\ \midrule
        E30 & \makecell[l]{$\begin{aligned}      &p \land \neg(p \lor q) \leftrightarrow p \land (\neg p \land \neg q) \leftrightarrow (p \land \neg p) \land \neg q \\
        &\leftrightarrow \text{False} \land \neg q \leftrightarrow \text{False}  \end{aligned}$} \\ \midrule
        E31 & \makecell[l]{$\begin{aligned}      &p \land \neg(p \lor q) \leftrightarrow (p \land \neg p) \land \neg q \leftrightarrow \text{False} \land \neg q \leftrightarrow \text{False}  \end{aligned}$} \\ \midrule
        E32 & \makecell[l]{$\begin{aligned}      &p \land \neg(p \lor q) \leftrightarrow p \land (\neg p \land \neg q) \leftrightarrow \text{False} \land \neg q \leftrightarrow \text{False}  \end{aligned}$} \\ \midrule
        E33 & \makecell[l]{$\begin{aligned}      &p \land \neg(p \lor q) \leftrightarrow p \land (\neg p \land \neg q) \leftrightarrow (p \land \neg p) \land \neg q \leftrightarrow \text{False}  \end{aligned}$} \\ 
        \bottomrule
    \end{tabular}
    }
    \caption{First Order Logic (FOL) Elimination Rules\protect\footnotemark}
    \label{tab:fol-eliminations-rules}
\end{table}

\newpage
\section{Complex FOL Expressions}
\label{sec:fol-complex-categories}
\begin{table}[htbp]
\centering
\small
\scalebox{0.78}{
\begin{tabular}{clccc}
\toprule
 & \makecell[c]{Symbolic Expression} & \makecell{Combination of\\FOL Rules}  & \makecell{Included\\in Training} \\
 \midrule
 \midrule
  C1 &\makecell[l]{$(((a\lor b)\rightarrow q)\land \neg q) \rightarrow (\neg a \land \neg b)$} & \makecell{MT + DM}& \xmark \\
 \midrule
 C2 &\makecell[l]{$(((a\land \neg b)\rightarrow q)\land \neg q) \rightarrow (\neg a \lor b)$} & \makecell{ MT + DM + DN}& \cmark \\
 \midrule
 C3 & \(
 \begin{array}{l}
     (p\rightarrow q), (q\rightarrow r), (s\rightarrow t), \\
     (\neg t \lor \neg r) \rightarrow (\neg p \lor \neg s)
 \end{array}
 \) & \makecell{TS + DD} & \cmark \\
 \midrule
 C4 & \(
 \begin{array}{l}
     (p\lor (q\land (a\lor b))) \leftrightarrow \\
     ((p\lor q)\land ((p\lor a)\lor b))
 \end{array}
 \) & \makecell{DS + AS} & \cmark\\
 \midrule
 C5 & \(
 \begin{array}{l}
     (p\land ((a \land b)\lor q \lor r)) \leftrightarrow \\
     (((p\land a)\land b)\lor (p\land q)\lor (p\lor r)) \leftrightarrow \\
     (((p\land a)\land b)\lor (p\land (q\lor r)))
 \end{array}
 \) & \makecell{DS + AS} & \cmark\\
 \midrule
 C6 & \(
 \begin{array}{l}
     ((p\land q\land r) \lor (a\land p\land b)\lor (c\land d\land e)) \leftrightarrow \\
     ((p\land ((q\land r)\lor(a \land b))) \lor (c\land d\land e))
 \end{array}
 \) & \makecell{DS + AS} & \xmark \\
 \midrule
 C7 & \(
 \begin{array}{l}
     (p\lor (q\land r \land (a\lor b) \land s)) \leftrightarrow \\
     ((p\lor q)\land (p\lor r)\land (p\lor a \lor b) \land (p\lor s))
 \end{array}
 \) & \makecell{DS + AS} & \cmark \\
 \midrule
 C8 & \(
 \begin{array}{l}
     (p\lor (q\land (p\lor b)\land r)) \leftrightarrow \\
     ((p\lor q)\land (p\lor b)\land (p\lor r)) \leftrightarrow \\
     (p\lor (q\land b\land r))
 \end{array}
 \) & \makecell{DS + AS + TT} & \cmark\\
 \midrule
 C9 & \(
 \begin{array}{l}
     \neg(p \lor (q\land(\neg a\lor b)\land \neg r)) \leftrightarrow \\
     \neg ((p\lor q)\land (p\lor \neg a\lor b)\land (p\lor \neg r)) \leftrightarrow \\
     (\neg (p\lor q) \lor \neg (p\lor \neg a \lor b) \lor \neg (p\lor \neg r)) \leftrightarrow \\
     ((\neg p \land \neg q)\lor(\neg p \land a \land \neg b)\lor(\neg p \land r)) \leftrightarrow \\
     (\neg p \land (\neg q \lor (a\land \neg b) \lor r))
 \end{array}
 \) & \makecell{DS + DM + DN} & \xmark \\
 \midrule
 C10 & \(
 \begin{array}{l}
     (\neg p\rightarrow q )\leftrightarrow (\neg q \rightarrow p)
 \end{array}
 \) & \makecell{TS + DN} & \cmark \\
 \midrule
 C11 & \(
 \begin{array}{l}
     (p\rightarrow \neg q )\leftrightarrow (q \rightarrow \neg p)
 \end{array}
 \) & \makecell{TS + DN} & \cmark \\
 \midrule
 C12 & \(
 \begin{array}{l}
     ((a\land b)\rightarrow q) \leftrightarrow (\neg q \rightarrow (\neg a \lor \neg b))
 \end{array}
 \) & \makecell{TS + DM} & \xmark \\
 \midrule
 C13 & \(
 \begin{array}{l}
     (p\rightarrow(\neg a\lor \neg b))\leftrightarrow ((a\land b)\rightarrow \neg p)
 \end{array}
 \) & \makecell{TS + DM} & \cmark \\
 \midrule
 C14 & \(
 \begin{array}{l}
     \neg ((a\lor b)\oplus c\oplus d)\leftrightarrow (\neg (a\lor b) \oplus \neg c \oplus \neg d)
 \end{array}
 \) & \makecell{DM + NX} & \cmark\\
 \midrule
 C15 & \(
 \begin{array}{l}
     \neg (c\oplus (\neg a\lor b)\oplus d)\leftrightarrow (\neg c\oplus (a\land \neg b)\oplus \neg d)
 \end{array}
 \) & \makecell{DM + NX} & \xmark \\
 \midrule
 C17 & \(
 \begin{array}{l}
     \neg(p \odot q\odot (a\lor \neg b)) \leftrightarrow (\neg p\odot \neg q \odot (\neg a \land b))
 \end{array}
 \) & \makecell{DM + NN} & \cmark\\
 \midrule
 C18 & \(
 \begin{array}{l}
     ((a\land b)\rightarrow q), ((a\land \neg c)\rightarrow s), (\neg q \lor \neg s) \rightarrow \\
     ((\neg a \lor \neg b)\lor (\neg a \lor c)) \rightarrow \\
     (\neg a \lor \neg b \lor c) \rightarrow \neg (a\land b \land \neg c)
 \end{array}
 \) & \makecell{DD + DN + DM + AS + TT} & \xmark \\
 \midrule
 C20 & \(
 \begin{array}{l}
     ((a\lor b)\rightarrow q), (r\rightarrow s), (a\lor b\lor r)\rightarrow (q\lor s)
 \end{array}
 \) & \makecell{CD + AS} & \cmark \\
 \midrule
 C21 & \(
 \begin{array}{l}
     (p\rightarrow (a\lor b)), (r\rightarrow s), (p\lor r)\rightarrow (a\lor (b\lor s))
 \end{array}
 \) & \makecell{CD + AS} & \cmark \\
 \midrule
 C22 & \(
 \begin{array}{l}
     (p\rightarrow q), ((a\lor \neg b)\rightarrow s), (p\lor \neg s) \rightarrow \\
     (q\lor (\neg a \land b)) \rightarrow ((q\lor \neg a)\land (q \lor b))
 \end{array}
 \) & \makecell{BD + DN + DM + DS} & \xmark \\
 \midrule
 C23 & \(
 \begin{array}{l}
     (p\rightarrow q), ((\neg a \land \neg b)\rightarrow s), (p\lor \neg s) \rightarrow \\
     q\lor \neg (\neg a\land \neg b) \rightarrow (q\lor a) \lor b
 \end{array}
 \) & \makecell{BD + DM + AS} & \cmark \\
\bottomrule
\end{tabular}
}
\captionsetup{skip=5pt}
\caption{Complex FOL Expressions\protect\footnotemark. (BD = Bidirectional Dilemma, CD = Constructive Dilemma, DD = Destructive Dliemma, MT = Modus Tollens, DM = De Morgan's, DN = Double Negation, DS = Distribution, AS = Association, TS = Transposition, TT = Tautology, NN = Negation of XNOR, NX = Negation of XOR)}
\label{table:fol-complex}
\end{table}

\newpage
\section{Training Data}
\label{sec:table_training_data}
\begin{table}[htbp]
\centering
\small
\scalebox{0.80}{
\begin{tabular}{ccc|ccc}
\toprule
  & \# Examples & \# Tokens & & \# Examples & \# Tokens\\
  \midrule
  \midrule
 BD &102.43K &778.57K     &DM &740.92K &34.11M\\
 CD &856.78K &71.70M     &Dist &268.76K &17.97M\\
 DD &806.15K &71.70M     &XOR$^*$ &17.46K &793.75K\\
 DS &321.31K &12.79M     &XNOR$^*$ &15.42K &668.07K\\
 HS &429.94K &23.20M     &XOR-XNOR$^*$ &14.49K &577.84K\\
 MP &237.80K &7.49M     & & &\\
 MT &285.19K &11.10M     & & &\\
 UI &123.00K &4.03M     & & &\\
 EG &9.10K &263.0K     & & &\\
 General/fol proof   &2.27M &286.12M     & & &\\
 \midrule
 C2 &94.36K &5.26M     &E0 &5.23K &83.92K\\
 C3 &108.56K &20.21M     &E1 &5.34K &141.32K\\
 C4 &102.73K &8.03M     &E2 &5.28K &147.74K\\
 C5 &107.64K &16.67M     &E3 &5.46K &88.60K\\
 C7 &109.03K &17.57M     &E4 &5.18K &82.91K\\
 C8 &97.90K &12.38M     &E5 &5.31K &146.08K\\
 C10 &102.43K &3.84M     &E6 &5.35K &148.27K\\
 C11 &86.92K &3.55M     &E7 &5.28K &84.61K\\
 C13 &93.22K &5.17M     &E8 &5.40K &188.11K\\
 C14 &109.20K &10.03M     &E9 &5.37K &194.38K\\
 C17 &108.53K &10.04M     &E10 &5.23K &140.78K\\
 C20 &109.10K &10.58M     &E11 &5.22K &139.83K\\
 C21 &109.37K &11.21M     &E12 &5.16K &134.20K\\
 C23 &109.45K &14.94M     &E13 &5.18K &135.07K\\
                & &     &E14 &5.60K &135.07K\\
                & &     &E15 &5.58K &720.92K\\
                & &     &E16 &5.56K &441.38K\\
                & &     &E17 &5.58K &569.09K\\
                & &     &E18 &5.61K &256.89K\\
                & &     &E19 &5.37K &291.75K\\
                & &     &E20 &5.21K &286.32K\\
                & &     &E21 &5.37K &352.57K\\
                & &     &E22 &5.61K &734.95K\\
                & &     &E23 &5.59K &470.14K\\
                & &     &E24 &5.63K &586.41K\\
                & &     &E25 &5.58K &703.58K\\
                & &     &E26 &5.57K &714.06K\\
                & &     &E27 &5.58K &501.50K\\
                & &     &E28 &5.57K &502.50K\\
                & &     &E29 &5.59K &625.27K\\
                & &     &E30 &5.59K &720.77K\\
                & &     &E31 &5.59K &499.31K\\
                & &     &E32 &5.59K &498.49K\\
                & &     &E33 &5.60K &648.88K\\
\textbf{Complex Total} & \textbf{1.45M} &\textbf{149.48M}     &\textbf{Eliminations Total} &\textbf{185.09K} &\textbf{12.24M}\\
 \midrule
 Random & 15.26M & 1.89B     & & &\\
 \midrule
 \midrule
 \textbf{Total} & \textbf{24.07M}& \textbf{2.67B}    & & &\\
 \bottomrule
\end{tabular}
}
\captionsetup{skip=5pt}
\caption{Full Breakdown of the Training Dataset.\protect\footnotemark The labels are consistent with the FOL types described in Table \ref{tab:fol-inference-rules}, Table \ref{tab:fol-basic}, Table \ref{tab:fol-eliminations-rules}, and Table \ref{table:fol-complex}. Note: not all basic properties (Table \ref{tab:fol-basic}) of FOL were included explicitly in generation. This is because we qualitatively saw that the massive random generation sufficiently and implicitly (and sometimes explicitly) captured the basic properties. \\ \\ *We explicitly included negations of XOR, negations of XNOR and the equivalences between XOR and XNOR.}
\label{table:train-data-breakdown}
\end{table}

\newpage
\section{Sanity Check \textsc{True}\,/\textsc{False} Experiment Plot}
\begin{figure*}[!h]
    \centering
    \includegraphics[width=\textwidth]{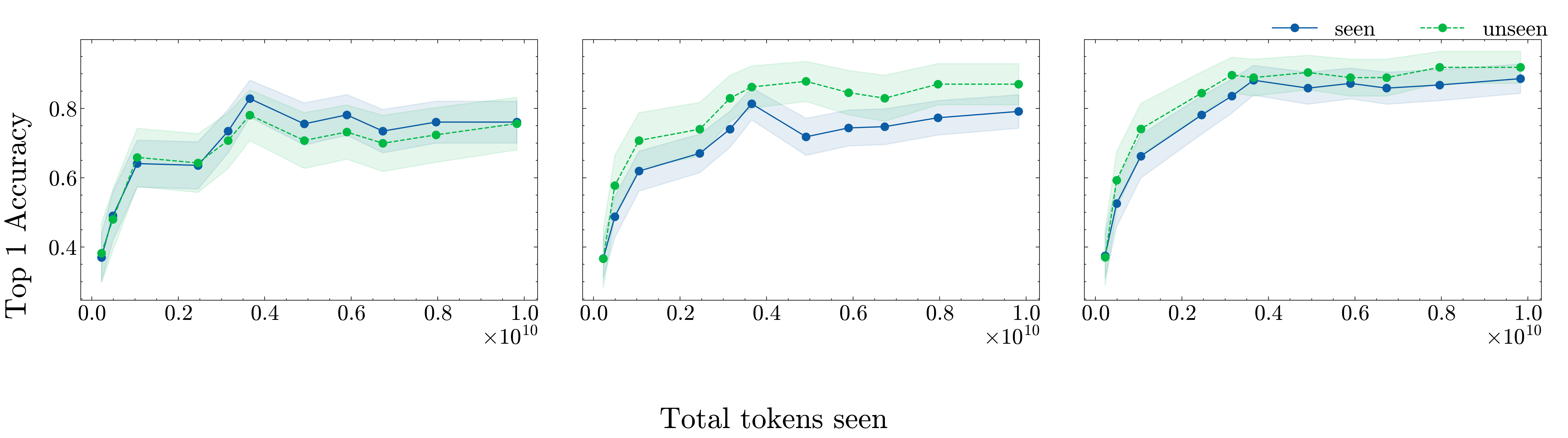}

                 \caption{Top-1 accuracy vs.\ cumulative tokens for three complexity tertiles (0--21, 22--32, $\geq$33). Shaded regions denote 95\% confidence interval. }
    \label{fig:top1_accuracy_complexity}
\end{figure*}

\section{Task 2 Prompts}
\begin{table}[h!]
\centering
\scriptsize
\begin{tabular}{p{0.46\columnwidth} p{0.46\columnwidth}}
\toprule
\textbf{1-Step Prompt} & \textbf{2-Step Prompt} \\
\midrule
\begin{minipage}[t]{\linewidth}\ttfamily
You are given a first-order logic equivalence chain where the last step is missing (replaced by <BLANK>).\\
Your task is to complete the chain by providing the final step that logically follows from the previous steps.\\
The chain uses \textbackslash Leftrightarrow\ to separate each step, and each step should be a valid logical expression.\\
Respond with ONLY the final step that should replace <BLANK>, with no explanation, no quotes, no extra text.\\
The final step should be a complete logical expression in parentheses.
\end{minipage}
&
\begin{minipage}[t]{\linewidth}\ttfamily
You are given a first-order logic equivalence chain where the last 2 steps are missing (replaced by <BLANK>).\\
Your task is to complete the chain by providing the final 2 steps that logically follow from the previous steps.\\
The chain uses \textbackslash Leftrightarrow\ to separate each step, and each step should be a valid logical expression.\\
Respond with ONLY the final 2 steps separated by \textbackslash Leftrightarrow\ that should replace the <BLANK>s, with no explanation, no quotes, no extra text.\\
Each step should be a complete logical expression in parentheses.
\end{minipage} \\
\bottomrule
\end{tabular}
\caption{System prompts for the 1-step and 2-step diagnostic tasks.}
\label{tab:diagnostic-prompts}
\end{table}

\section{Task 2 Example}
\begin{table}[h!]
\centering
\small
\setlength{\tabcolsep}{3pt}
\renewcommand{\arraystretch}{1.15}
\begin{tabularx}{\columnwidth}{lX}
\toprule
\textbf{Examples} & \textbf{Expressions and Evaluation Results} \\
\midrule
\textbf{\makecell[l]{Step-2 correct,\\chain wrong}}
 &
\textbf{Answer:} $P \Leftrightarrow \neg P \Leftrightarrow \text{False}$ \newline
\textbf{Prediction:} $P \Leftrightarrow Q \Leftrightarrow \text{False}$ \newline
$acc_{\text{2step\_step2}} = 1$ (both end with False) \newline
$acc_{\text{2step\_chain}} = 0$ \; [$(P \Leftrightarrow \neg P) \Leftrightarrow \text{False} \neq (P \Leftrightarrow Q) \Leftrightarrow \text{False}$] \\[3pt]
\textbf{\makecell[l]{Different Step-2,\\chain still correct}} &
\textbf{Answer:} $P \Leftrightarrow \neg P \Leftrightarrow \text{False}$ \newline
\textbf{Prediction:} $P \Leftrightarrow (P \land \neg P) \Leftrightarrow \text{False}$ \newline
$acc_{\text{2step\_step2}} = 1$ (both end with False) \newline
$acc_{\text{2step\_chain}} = 1$ \; [$(P \Leftrightarrow \neg P) \Leftrightarrow \text{False} \equiv (P \Leftrightarrow (P \land \neg P)) \Leftrightarrow \text{False}$] \\
\bottomrule
\end{tabularx}
\caption{Examples illustrating different evaluation outcomes for the \textit{step completion} task.}
\label{tab:step_completion_examples}
\end{table}

\end{document}